%% file: main.tex
\documentclass[10pt,twocolumn,letterpaper]{article}

\usepackage[pagenumbers]{iccv} 
\usepackage{multirow}
\usepackage{amssymb}
\usepackage{array}
\usepackage{makecell} 
\usepackage{colortbl}  

\definecolor{iccvblue}{rgb}{0.21,0.49,0.74}
\usepackage[pagebackref,breaklinks,colorlinks,allcolors=iccvblue]{hyperref}

\def\paperID{6010} 
\def\confName{ICCV}
\def\confYear{2025}

\title{DEPTHOR: Depth Enhancement from a Practical Light-Weight \\dToF Sensor and RGB Image}

\author{
Jijun Xiang$^{1}$,
~~Xuan Zhu$^{1}$,
~~Xianqi Wang$^{1}$,\\
~~Yu Wang$^{2}$,
~~Hong Zhang$^{2}$,
~~Fei Guo$^{2}$,
~~Xin Yang$^{1}$\footnotemark[2]\\
[2mm]
$^1$~Huazhong University of Science and Technology \quad $^2$~Honor Device Co., Ltd \\
{\tt\small \{jijunx, xuanzhu, xianqiw, xinyang2014\}@hust.edu.cn}\\
{\tt\small \{wangyu24, zhanghong70, guofei2\}@honor.com}
}

\begin{document}
\maketitle
\input{sec/0_abstract}
\renewcommand{\thefootnote}{\fnsymbol{footnote}}
\footnotetext[2]{Corresponding author.}
\input{sec/1_intro}
\input{sec/2_formatting}
\input{sec/3_finalcopy}

\input{sec/4_experiment}

\input{sec/5_conclusion}
{
    \small
    \bibliographystyle{ieeenat_fullname}
    \bibliography{main}
}

\input{sec/X_suppl}
\end{document}

%% file: sec/0_abstract.tex
\begin{abstract}
Depth enhancement, which uses RGB images as guidance to convert raw signals from dToF into high-precision, dense depth maps, is a critical task in computer vision. Although existing super-resolution-based methods show promising results on public datasets, they often rely on idealized assumptions like accurate region correspondences and reliable dToF inputs, overlooking calibration errors that cause misalignment and anomaly signals inherent to dToF imaging, limiting real-world applicability. To address these challenges, we propose a novel completion-based method, named DEPTHOR, featuring advances in both the training strategy and model architecture. First, we propose a method to simulate real-world dToF data from the accurate ground truth in synthetic datasets to enable noise-robust training. Second, we design a novel network that incorporates monocular depth estimation (MDE), leveraging global depth relationships and contextual information to improve prediction in challenging regions. On the ZJU-L5 dataset, our training strategy significantly enhances depth completion models, achieving results comparable to depth super-resolution methods, while our model achieves state-of-the-art results, improving Rel and RMSE by 27\% and 18\%, respectively. On a more challenging set of dToF samples we collected, our method outperforms SOTA methods on preliminary stereo-based GT, improving Rel and RMSE by 23\% and 22\%, respectively. Our Code is available at \url{https://github.com/ShadowBbBb/Depthor}
\end{abstract}

%% file: sec/1_intro.tex
\section{Introduction}
\label{sec:intro}
The direct Time-of-Flight (dToF) sensor is a depth sensor that measures depth by calculating the time it takes for emitted light pulses to reflect off objects and return to the sensor. With advantages like miniaturization and low power consumption, dToF sensors are widely deployed on mobile devices for applications like autofocus and obstacle detection \cite{deltar,autofocus}. However, the depth information provided by dToF sensors is typically too coarse for high-precision tasks like 3D reconstruction \cite{Sr-lio, Sr-livo} and SLAM \cite{liu2023multi, yang2019fast, dai2017bundlefusion, izadi2011kinectfusion, SDV-LOAM}. A common solution is using depth enhancement methods to produce dense, high-resolution depth maps from raw sensor data, using RGB images as guidance.

Based on the sensor's data format, these methods fall into two categories: depth completion and depth super-resolution. In depth completion, the sensor generates a high-resolution depth map where valid depth points are sparsely distributed. The algorithm then propagates these sparse measurements to reconstruct dense depth through geometric reasoning guided by RGB context. Conversely, in depth super-resolution, the sensor returns a low-resolution dense depth map where each element corresponds to a local image region. The algorithm subsequently upsamples the depth map to match RGB resolution by recovering high-frequency details through cross-modal guidance.

Existing dToF enhancement approaches \cite{deltar,dvsr,ding2024cfpnet} are typically designed for depth super-resolution, which are generally based on two assumptions: \emph{(1) ideal calibration exists between the RGB camera and dToF sensor,} and \emph{(2) the dToF sensor operates reliably}. However, these assumptions often fall short in real-world environments. Through a systematic analysis of the RGB-dToF samples collected from a mobile phone, we found that calibration errors between devices are inevitable and may increase over time, leading to misalignment and conflicts in region correspondences, with a maximum deviation of up to two dToF pixels. In addition, the imaging principle of the dToF sensor results in signal loss or anomalous values in specific regions.

To address these challenges, we project dToF signals into a high-resolution sparse depth map using device parameters, redefining the problem within the scope of depth completion rather than depth super-resolution. The key motivation behind this projection is that, after transformation, both calibration errors and signal anomalies manifest as depth point inconsistencies at global or local scales. This reformulation allows us to focus solely on improving the robustness of the depth completion model against anomalous depth measurements, eliminating the need for complex region correspondences and thereby relaxing the restrictive assumptions of previous methods.

For this depth completion task with noisy input, we propose a noise-robust training strategy consisting of two key aspects: training dataset selection and dToF simulation, whose core idea is to simulate real-world dToF inputs during training while ensuring accurate supervision. While high-precision sensor-acquired datasets are commonly used in previous methods, their ground truth often lacks high-frequency details and exhibits similar anomaly patterns to dToF, limiting their effectiveness for training. In contrast, synthetic datasets offer more precise and detailed supervision, making them a preferable choice. To further improve alignment with real-world dToF characteristics, we introduce a dToF simulation method that accounts for four key aspects: overall distribution, specific abnormal regions, calibration errors, and random anomalies.

Although our training strategy effectively improves models' prediction on real-world dToF data, correcting erroneous dToF signals or distinguishing general empty regions from signal loss remains an ill-posed problem for many methods, as they primarily focus on propagating existing depth measurements. Thus, we propose a simple yet effective depth completion network that integrates the monocular depth estimation (MDE) model, leveraging its global depth relationships and contextual information to improve predictions in challenging regions. Specifically, our model consists of two stages: multimodal fusion and refinement. First, we employ an encoder-decoder to extract and aggregate RGB and depth features along with the relative depth map, generating an coarse prediction. Then, we fuse the MDE feature with the decoder feature, compute mixed affinity, and further refine the initial depth map.

Our contributions are as follows:
\begin{itemize}
    \item[$\bullet$] We conduct a comprehensive analysis of real-world dToF data and propose a noise-robust training strategy with a novel dToF simulation method on synthetic datasets.
    
    \item[$\bullet$] We design a depth completion network that effectively integrates MDE model at multiple stages to enhance predictions in ill-posed regions inherent to dToF imaging.

    \item[$\bullet$] On the ZJU-L5 dataset, our training strategy enhances depth completion models, achieving results comparable to depth super-resolution methods. Meanwhile, our model surpasses all types of state-of-the-art methods, improving Rel and RMSE by 27\% and 18\%, respectively.
    
    \item[$\bullet$] On a more challenging set of mobile phone-based dToF samples we collected, our model outperforms SOTA models on preliminary GT generated by stereo matching, improving Rel and RMSE by 23\% and 22\%.
\end{itemize}

%% file: sec/2_formatting.tex
\section{Related Work}
\label{sec:related}
\noindent \textbf{Depth Completion.} Conventional 2D depth completion methods can be broadly categorized into encoder-decoder approaches \cite{acmnet,completionformer,penet} and affinity propagation approaches \cite{wang2023lrru,cspn++,cspn}, with most evaluations conducted on real datasets \cite{kitti, nyuv2, arkitscenes}. Recent methods \cite{bpnet,tpvd,decotr} utilize camera parameters to project depth points into 3D space, enhancing accuracy by leveraging spatial information. Additionally, some methods focus on aspects such as point sparsity 
\cite{sparsity0,sparsity1,sparsity2,steeredmarigold} and cross-dataset generalization 
\cite{testtime, omnidc, depthlab} to further improve practicality.

However, assessing the applicability of existing methods to real-world dToF data remains challenging due to two key limitations. First, many methods rely on idealized data simulations. For instance, on the NYUv2 dataset, 500 perfectly accurate depth points are randomly sampled from GT, causing evaluation metrics to focus on depth propagation rather than robustness to sensor noise in real-world scenarios. Second, the unique characteristics of dToF data are rarely considered. To the best of our knowledge, only methods from the MIPI competition \cite{mipi2022,mipi2023,emdc} have attempted to simulate the uniform distribution of dToF data using grid sampling. Therefore, many designs that are considered effective in other depth modalities are not suitable for dToF.

\noindent \textbf{Depth Super-resolution.} For dToF data, Deltar \cite{deltar} proposed a dual-branch depth super-resolution network that utilizes PointNet to extract dToF features and employs a transformer-based fusion module to integrate RGB and depth information. Building upon this, CFPNet \cite{ding2024cfpnet} addressed the limited FoV coverage of dToF sensors by incorporating large convolution kernels and cross-attention mechanisms to enhance predictions in border regions. DVSR \cite{dvsr} specifically addresses depth super-resolution in dToF video sequences, using optical flow and deformable convolutions to aggregate multi-frame information, thereby enhancing prediction consistency. 

The dToF simulation in these approaches typically begin by computing the depth histogram within a given region of the ground truth, followed by processing the histogram based on the characteristics of the target sensor(\eg mean, peak, variance, rebin histograms). Among these, DVSR accounts for signal loss in low-reflectance regions by estimating the probability of missing depth measurements based on the mean RGB value. 

However, existing methods typically rely on accurate RGB-dToF correspondence. For example, the ZJU-L5 dataset provides the coordinates of the RGB region corresponding to each dToF signal, and both Deltar and CFPNet leverage these coordinates to guide their feature aggregation modules. DVSR assumes that the dToF data is uniformly distributed, dividing a 480 $\times$ 640 image into 30 $\times$ 40 patches and directly simulating the dToF from depth GT corresponding to each patch. When real-world devices fail to provide accurate correspondences, the performance of these methods deteriorates significantly.

\noindent \textbf{Monocular Depth Estimation.} These methods predict the depth of each pixel in the input RGB image to obtain a dense depth map. Early methods trained on each dataset predict metric depth, but due to the inherent lack of depth scale information, these methods have poor generalization across different datasets. Ranftl \etal~\cite{midas} introduced an affine-invariant loss, which predicts the inverse depth 1/d, making the model focus more on relative distance relationships rather than absolute depth values. This mitigates the impact of scale shifts between different datasets and further enhances the model's generalization.

Recent models \cite{depthanythingv1,depthanythingv2,metric3d,metric3dv2,marigold} have significantly advanced this field with methods like pseudo-label generation, diffusion model priors, and additional normal supervision. Among these, Ke \etal~\cite{marigold} achieves high detail in depth image outputs; however, its reliance on a diffusion model results in long inference times, which conflicts with many dToF application scenarios. Hu \etal~\cite{metric3dv2}, by introducing a normalized camera model and additional normal supervision, enables the model to output scaled depth with some generalization. Depth Anything \cite{depthanythingv1,depthanythingv2}, on the other hand, outputs inverse depth and, through techniques like pseudo-label generation and teacher-student models, provides both generalization and detail performance. Since dToF sensors already offer depth scale information but lack detail, we select Depth Anything V2 \cite{depthanythingv2} as the pre-trained monocular depth estimation model for our subsequent experiments.

%% file: sec/3_finalcopy.tex
\section{The Proposed Method}
\label{sec:method}

\subsection{Training Strategy with dToF Data Simulation}
\label{sec:strategy}
\begin{figure}[htbp]
    \centering
    \includegraphics[width=1\linewidth]{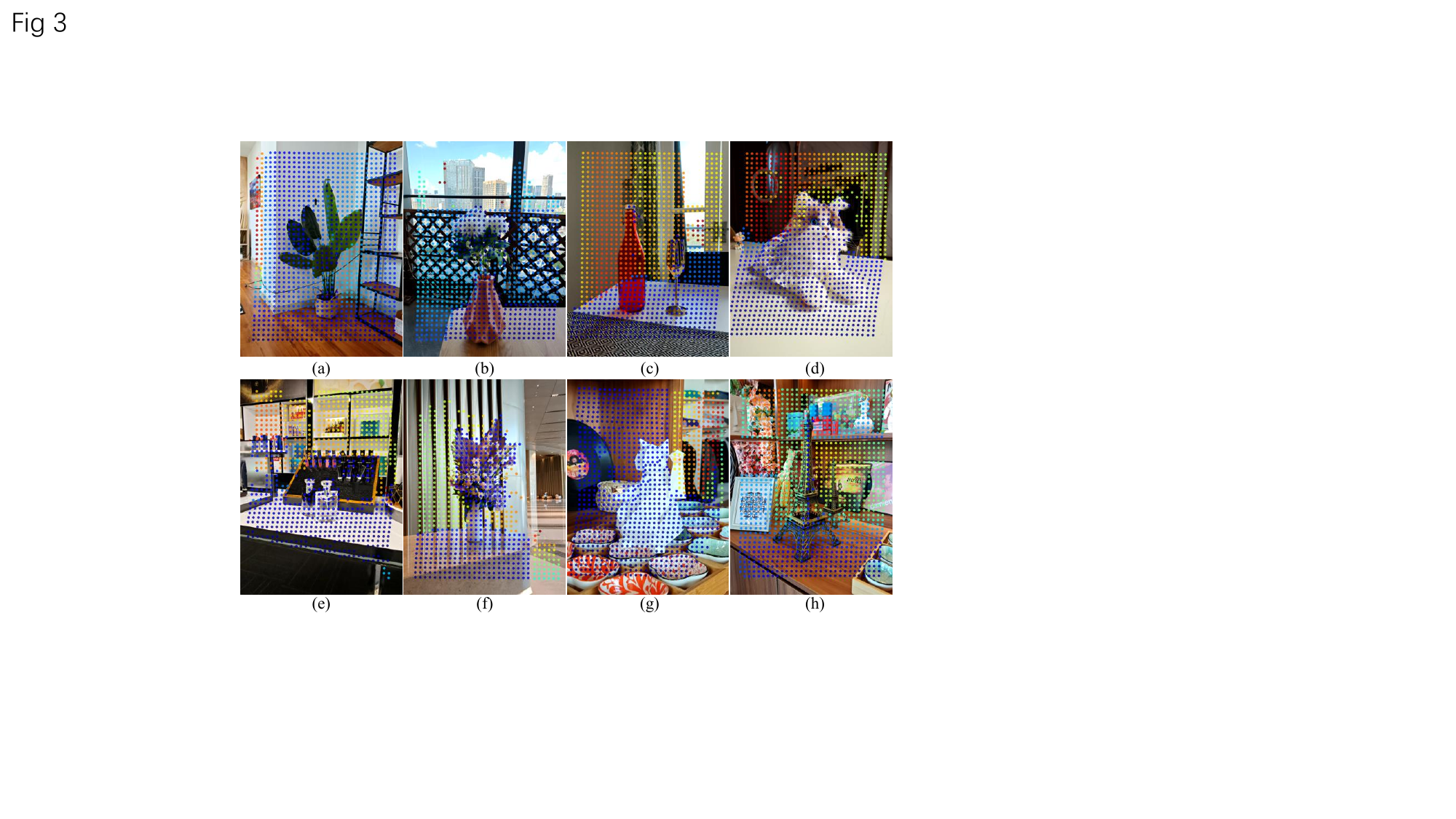}
    \caption{Ideal and anomalous samples in real-world dToF data.}
    \label{fig:abnormal}
    \vspace{-1em} 
\end{figure}

We collected a set of RGB-dToF samples using an Honor Magic6 Ultra to analyze the distribution characteristics and potential anomalies of real-world dToF data, with resolutions of $912\times684$ and $40\times30$, respectively. Readers interested in dToF imaging are referred supplementary material for more details. \Cref{fig:abnormal}a shows an RGB-dToF sample obtained under ideal conditions.

Using the intrinsic and extrinsic parameters of both the dToF sensor and the RGB camera, along with calibration transformation matrices, we project dToF data into a high-resolution, uniformly distributed sparse depth map. For this depth map, we design our simulation method based on the following four key aspects:

\begin{figure*}[htbp]
	\centering
	\includegraphics[width=0.95\textwidth]{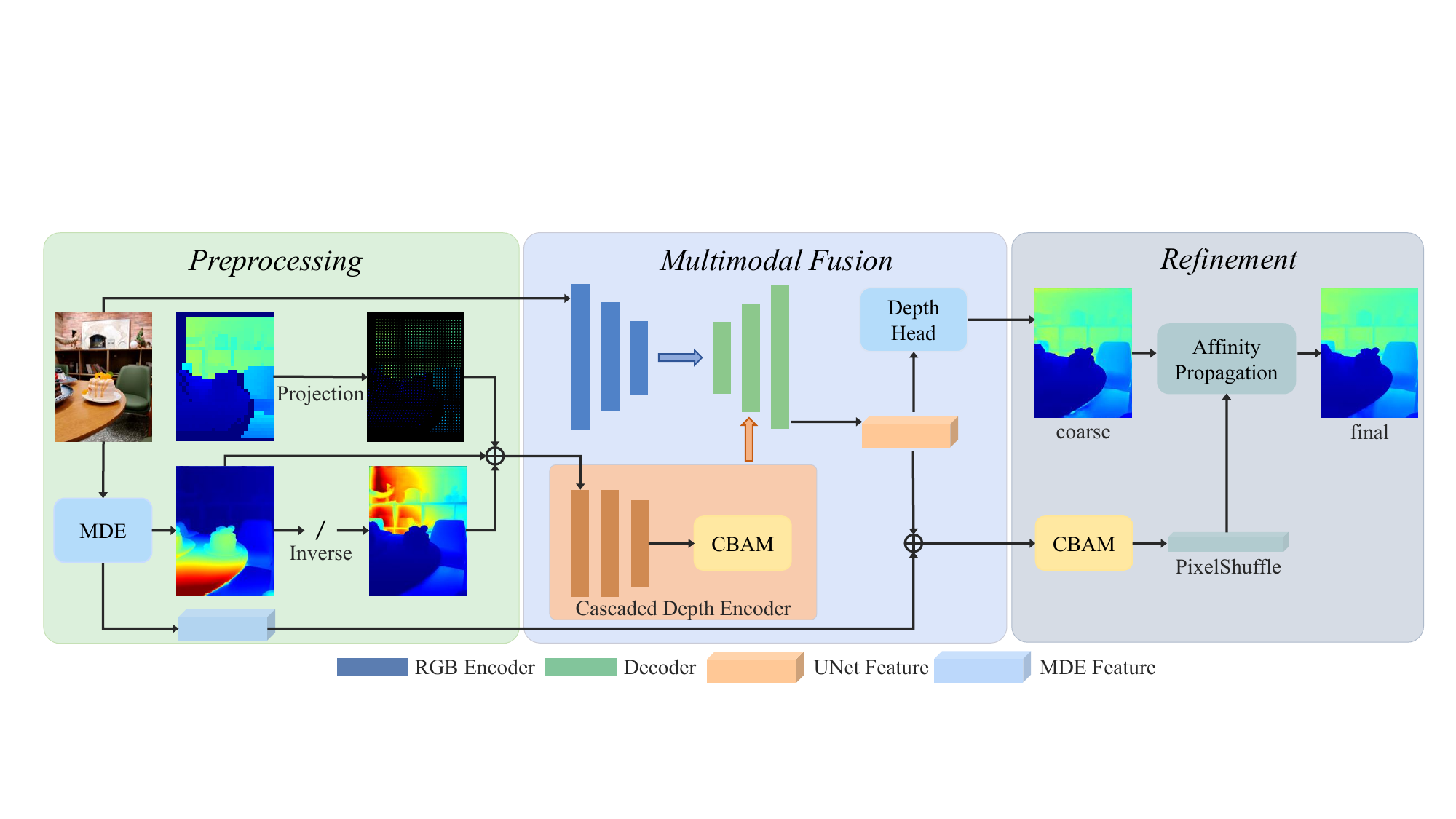}
	\caption{\textbf{Overview of our model:} We first project dToF signals into sparse depth points, and use a pre-trained MDE model to generate inverse and relative depth maps. In multimodal fusion, we employ a simple encoder-decoder structure to obtain a coarse estimation. In refinement, we update the initial depth map using mixed affinity propagation.}
	\label{fig:structure}
	\vspace{-1em} 
\end{figure*}

\noindent{\textbf{Overall Distribution.}} Due to the substantial difference in FoV between dToF sensor and RGB camera, the captured depth data does not cover the entire RGB image. Moreover, the high-resolution sparse depth points are inherently imprecise, as they theoretically correspond to peak values within a defined iFoV (\cref{fig:abnormal}h). Thus, we perform random translations and rotations on the depth GT within the iFoV range, followed by approximately uniform sampling within the roughly defined FoV.

\noindent{\textbf{Specific Abnormal Regions.}} We also address abnormal conditions in specific areas inherent to dToF imaging.

\begin{enumerate}
    \item \textit{Non-Lambertian surfaces} (\cref{fig:abnormal}b, \cref{fig:abnormal}c): Photons may pass through objects, leading to signal loss or returning depth values from a farther distance. We use diffuse reflection intensity to identify non-Lambertian regions and randomly determine the type of anomaly.
    \item \textit{Low-reflectivity areas} (\cref{fig:abnormal}e): In low-light environments or on dark surfaces, photons are absorbed rather than reflected, leading to signal loss. We convert RGB to HSV space and assign a probability of signal loss to points with low brightness values in the V channel.
    \item \textit{Long-distance regions} (\cref{fig:abnormal}f): At greater distances, photons are more susceptible to environmental noise and may be lost entirely if they exceed the device's maximum reception time. The theoretical maximum detection range of our dToF sensor is 8.1 meters, but signals beyond 6 meters are frequently lost in practical use.
\end{enumerate}

\noindent{\textbf{Random Anomalies.}} To enhance the model's robustness to random anomalies (\cref{fig:abnormal}g), we introduced approximately 5\% noise points and 5\% blank points. The depth values of the noise points were randomly assigned within the theoretical detection range.

\noindent{\textbf{Calibration Errors.}} Calibration errors manifest as regional shifts after 
projection. Specifically, foreground points generally project with high precision, while background points often experience a noticeable shift (\cref{fig:abnormal}d). Therefore, we select a percentile from GT as the threshold, treat points above it as background, and apply a random shift (within 0–2 dToF pixels).

Using this simulation method, we obtain training dToF data from synthetic datasets that closely resemble real-world distributions. Supervising this input with accurate ground truth encourages the model to learn robust features under imperfect data conditions, thereby better adapting to real-world conditions and partially mitigating issues inherent to dToF imaging, as shown in \cref{fig:intro}.

\subsection{Depth Completion Model Integrating MDE}
\label{sec:model}
We formulate the problem as: given a projected sparse depth map $S$, the corresponding RGB image $I$, and the inverse depth map ${D}_{inv}$ and features ${F}_{mde}$ output by the MDE model (based on $I$), the goal of the depth completion model is to predict a dense depth map $D$.

As shown in \cref{fig:structure}, our proposed network consists of two stages: multi-modal fusion and refinement. The first stage outputs an initial depth map at half resolution, while the second stage refines it to produce the precise full-resolution depth prediction using affinity propagation.

\noindent \textbf{Multimodal Fusion.} In this stage, we implement an encoder-decoder network. The encoder extracts multi-resolution features from image and depth separately, which are subsequently fused in the decoder. The fused feature ${F}_{unet}$ is then passed through a depth head to produce an initial depth estimation.


We employed the network from BPNet \cite{bpnet} as the RGB encoder, progressively downsampling the RGB image and generating feature maps $F_{img}$ at resolutions ranging from 1/2 to 1/32. We modified its architecture and feature dimensions to reduce computational cost and parameters.

Since distant regions in ${D}_{inv}$ are numerically compressed toward zero, we introduce the relative depth map ${{D}_{rel}}=1/({{D}_{inv}}+\varepsilon )$ to emphasize structural details in these areas.
The inputs of depth encoder include \(\{D_{rel}, D_{inv}, S\}\), while ${D}_{rel}$ and ${D}_{inv}$ are normalized, $S$ remains unnormalized to retain absolute scale information. With these simple design choices, the depth encoder maintains a stable balance between near and far regions, as well as between relative and absolute depth.

To effectively extract depth features $F_{dep}$, we first apply a combination of convolution layers, including large-kernel dilated convolution to enhance the perception of scale information in $S$ and small-kernel downsampling convolution to capture high-frequency details in $D_{rel}$ and $D_{inv}$. Then, we feed the output feature into a CBAM module \cite{cbam}, where spatial and channel attention mechanisms are employed for aggregation.

In the decoder, we progressively fuse RGB and depth features through convolution and upsampling layers, ultimately producing the decoder feature $F_{unet}$.

Following the design of \cite{deltar,adabins}, we input $F_{unet}$ into a depth head. Rather than directly regressing depth values, the depth head generates a set of $N$ non-uniformly normalized depth bins $b$ for each image, along with weighting coefficients $k_i$ for each pixel corresponding to $b_i$. After restoring the depth bins to metric depth using hyperparameters and computing each bin's center $c_i$, the initial depth is computed using the following formula:

\begin{align}
d=\sum\limits_{i=1}^{N}{{{k}_{i}}{{c}_{i}}} \label{eq:depthhead}
\end{align}

Since computing the depth map requires generating $N$-dimensional features for each pixel, we set 
$N$ to 128 to balance computational cost and accuracy. Additionally, the initial depth map is predicted at half resolution.

\noindent \textbf{Refinement.} The initial depth map often contains various anomalies, such as artifacts in regions without depth measurement coverage and residual erroneous signals in certain areas, which is particularly pronounced when the MDE model produces sharp numerical changes. To address this, we deploy an affinity propagation module, based on CSPN++\cite{cspn++}, to further refine the initial depth map.

Unlike previous methods that compute affinity using single-modality features from the decoder, we jointly compute affinity, since the rich semantic information in ${F}_{mde}$ helps correct errors in ${F}_{unet}$ caused by inaccurate depth signals. Conversely, incorporating ${F}_{unet}$ mitigates the resolution discrepancies introduced by the Transformer architecture and the lack of scale information in ${F}_{mde}$.

We begin by interpolating ${F}_{mde}$ to align with the half-resolution of ${F}_{unet}$. We then concatenate ${F}_{unet}$ and ${F}_{mde}$, feeding this combined feature into a CBAM module and a PixelShuffle \cite{pixelshuffle} layer to upsample to the full resolution. Using this merged feature ${F}_{cspn}$, we calculate mixed affinity ${\omega}_k$ corresponding to kernel $[3,5,7]$. 

During the propagation, the update process of pixel $i$ under affinity kernel $k$ at the 
$t$-th iteration is formulated in \eqref{eq:propagation}. 

\begin{align}
\hat{D}_{i,k,t} = \omega_{i,k} \hat{D}_{i,t-1} + \sum_{j \in \mathbb{N}_k(i)} \omega_{j,k} \hat{D}_{j,t-1} \label{eq:propagation}
\end{align}

Following BPNet\cite{bpnet}, we aggregate the outputs across different iterations and affinity kernels using two normalized weights produced by a convolution and softmax layer, as described in \eqref{eq:aggregation}, where $t\in \{0,T/2,T\}$.

\begin{align}
D = \sum_{t \in T} \tau_t \sum_{k \in \mathcal{K}} \sigma_k \hat{D}_{k,t} \label{eq:aggregation}
\end{align}

Conventional affinity propagation modules typically embed sparse depth measurements at each iteration, directly assigning the original sparse depth values to the updated depth map. However, since dToF points are not entirely accurate, we remove this setting.

\subsection{Implementation Details}
\label{sec:detail}
We employ a scaled affine-invariant loss for supervision following \cite{deltar}, the expression of which is as follows:

\begin{align}
L=\alpha \sqrt{\frac{1}{T}\sum\limits_{i}{g_{i}^{2}-}\frac{\lambda }{{{T}^{2}}}{{(\sum\limits_{i}{{{g}_{i}}})}^{2}}} \label{eq:loss}
\end{align}

Where ${{g}_{i}}=\log {{\tilde{d}}_{i}}-\log {{d}_{i}}$, ${{\tilde{d}}_{i}}$, ${{d}_{i}}$, represent the predicted values and ground truth for valid pixel points, respectively, and in all experiments, $\alpha$=10, $\lambda$=0.85. We calculate the loss only for pixels within the sensor's theoretical detection range. The detailed training settings are provided in the supplementary material.

%% file: sec/4_experiment.tex
\section{Experiments}
\label{sec:experiment}
\subsection{Datasets \& Evaluation Metrics}
\label{sec:datasetmetric}
\noindent{\textbf{Hypersim dataset for training.}} We trained our model on the Hypersim \cite{hypersim} dataset, as it provides precise ground truth and extensive labels, with 59,544 frames for training and 7,386 frames for testing. For different dToF sensors, we modified the simulation method to ensure a similar distribution between the training and testing data. Please refer to the supplementary material for more details.

\noindent{\textbf{ZJU-L5 dataset for testing.}} Deltar\cite{deltar} employs the ST VL53L5CX (L5) and the Intel RealSense 435i to capture raw dToF data and ground truth. The dToF and image resolutions are $8\times8$ and $480\times640$, respectively. In our method, we utilize the provided iFoV to convert each dToF signal into a sparse depth point at the center of its corresponding region, without using the variance information.

\noindent{\textbf{Real-world samples we collected for testing.}} We collected RGB-dToF samples on an Honor Magic6 Ultra and projected dToF using its internal parameters. We use the results from an SOTA stereo matching method \cite{monster} as a preliminary ground truth for evaluation.  The detailed pipeline is introduced in supplementary material.

\begin{figure}[htbp]
	\centering
	\includegraphics[width=\columnwidth]{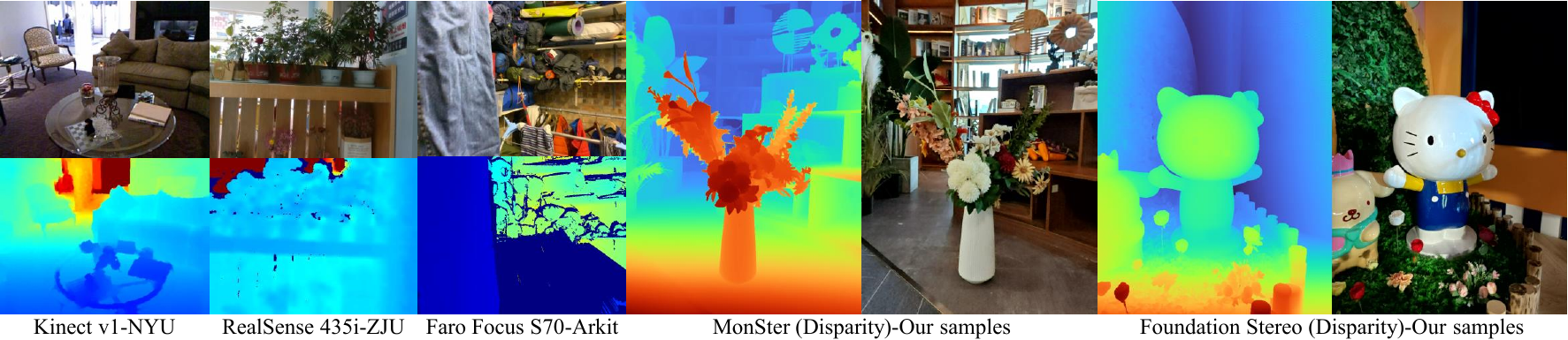}
	\vspace{-14pt}
	\caption{Ground truth from various sensors and stereo methods}
	\vspace{-8pt}
	\label{fig:rebuttal_gt}
\end{figure}

We adopt a high-quality stereo matching pipeline using the main and ultra-wide cameras on the mobile phone to get ground truth, manually filter failed samples, and mask noisy regions. Lens distortion and baseline mismatch may slightly affect the epipolar geometry, leading to a global shift. While not perfect, we believe the SOTA stereo methods \cite{igev++, fastacc, monster, foundationstereo} to be more effective than common sensors, especially in complex regions targeted in our work (\cref{fig:rebuttal_gt}), the results in the main paper are from MonSter \cite{monster}.

\noindent{\textbf{Evaluation Metrics}} We reported standard metrics including ${\delta}_{i}$, Rel, RMSE, $\text{log}_{10}$. To further evaluate performance at boundaries, we also reported edge-weighted mean absolute error (EWMAE) \cite{mipi2023,ewmae}, which assigns greater weight to pixels with larger gradients when calculating MAE. The details are introduced in the supplementary material.

\subsection{Effectiveness of Our Training Strategy.}
\label{sec:analysisstrategy}
We trained our model and a lightweight PENet (denoted as PENet*) using different simulation methods on Hypersim and evaluated on the ZJU-L5. For PENet*, we retain the original design but reduce the number of layers and channels to accelerate training. As a result, the parameters and FLOPs are reduced from 131M / 592G to 48M / 110G.

As shown in \cref{tab:comparison}, our method significantly improves the performance of both models. Notably, PENet* even outperforms the SOTA super-resolution method CFPNet on several metrics, demonstrating that our simulation strategy effectively narrows the gap between depth completion and super-resolution, without relying on precise dToF-RGB region correspondences. We further analyze the impact of training datasets in \cref{sec:ablation}.

\begin{table}[ht]
    \vspace{-6pt}
    \centering
    \renewcommand{\arraystretch}{1}
    \setlength{\tabcolsep}{5pt}
    \resizebox{\linewidth}{!}{
    \begin{tabular}{lcccccc}
        \toprule
        Model  & Simulation & $\delta_1$ & $\delta_2$  & Rel & RMSE & $\log_{10}$\\
        \midrule
        CFPNet & Deltar & 0.883 & \underline{0.949} & 0.103 & \underline{0.431} & 0.047\\
        PENet  & Deltar & 0.807 & 0.914 & 0.161 & 0.498 & 0.065\\
        \midrule
        PENet* & Deltar & 0.815 & - & 0.152 & 0.510 & - \\
        PENet* & MIPI   & 0.865 & 0.929 & 0.118 & 0.493 & 0.061\\
        PENet* & Ours   & \underline{0.889} & \underline{0.949} & \underline{0.093} & 0.447 & \underline{0.046}\\
        \midrule
        Ours   & Deltar & 0.804 & 0.883 & 0.164 & 0.562 & 0.097 \\
        Ours   & MIPI   & 0.853 & 0.909 & 0.123 & 0.511 & 0.089 \\
        Ours   & Ours   & \textbf{0.933} & \textbf{0.972} & \textbf{0.075} & \textbf{0.350} & \textbf{0.034}\\
        \bottomrule
    \end{tabular}
    }
    \vspace{-6pt}
    \caption{\textbf{Performance under different simulation methods.} The results in the second row are reported by Deltar\cite{deltar}.}
    \label{tab:comparison}
    \vspace{-8pt}
\end{table}

\subsection{Comparison with SOTAs}
\label{sec:comparison}
In addition to referencing results from published papers, we conducted additional experiments to ensure a comprehensive comparison. \textbf{For monocular depth estimation}, we separately evaluated the \textbf{M}etric-\textbf{L}arge (fine-tuned on Hypersim) and \textbf{R}elative-\textbf{S}mall (used in our method) versions of Depth Anything V2 (indicated as DAv2 -ML and DAv2 -RS). The former was tested directly, while for the latter, we fitted its output using dToF measurements. \textbf{For depth completion}, we conducted two types of experiments: (1) retraining existing methods with our strategy, including the PENet*, the SOTA 2D (CompletionFormer) and 3D (BPNet) method on conventional benchmark; (2) testing the latest generalizable model OMNI-DC, which incorporates multiple depth modalities and varying levels of sparsity during training.

\begin{table}[ht]
    \centering
    \renewcommand{\arraystretch}{1.2} 
    \setlength{\tabcolsep}{6pt} 
    \vspace{-5pt}
    \resizebox{\linewidth}{!}{ 
        \begin{tabular}{lcc|ccccc}
            \toprule
            Method & Type & Pub & $\delta_1$ & $\delta_2$ & Rel & RMSE & $\log_{10}$ \\
            \midrule
            BTS\cite{bts} & MDE & arXiv19 & 0.739 & 0.914 & 0.174 & 0.523 & 0.079 \\
            AdaBins\cite{adabins} & MDE & CVPR21 & 0.770 & 0.926 & 0.160 & 0.494 & 0.073 \\
            PnP-Depth\cite{pnpdepth} & DS & ICRA19 & 0.805 & 0.904 & 0.144 & 0.560 & 0.068 \\
            PrDepth\cite{xia2020generating} & DS & CVPR20 & 0.800 & 0.926 & 0.151 & 0.460 & 0.063 \\
            PENet\cite{penet} & DC & ICRA21 & 0.807 & 0.914 & 0.161 & 0.498 & 0.065 \\
            Deltar\cite{penet} & DS & ECCV22 & 0.853 & 0.941 & 0.123 & 0.436 & 0.051 \\
            CFPNet\cite{ding2024cfpnet} & DS & 3DV25 & 0.883 & \underline{0.949} & 0.103 & \underline{0.431} & 0.047 \\
            \midrule
            PENet*\cite{penet} & DC & ICRA21 & \underline{0.889} & \underline{0.949} & \underline{0.093} & 0.447 & \underline{0.046}\\
            BPNet\cite{bpnet} & DC & CVPR24 & - & - & - & 0.671 & - \\
            CFormer\cite{completionformer} & DC & CVPR23 & 0.873 & 0.938 & 0.103 & 0.480 & 0.053 \\
            OMNI-DC\cite{omnidc} & DC & arXiv24 & 0.871 & 0.933 & 0.099 & 0.502 & 0.053 \\
            DAv2 -ML \cite{depthanythingv2} & MDE & Neurips24 & 0.703 & 0.905 & 0.220 & 0.467 & 0.083 \\
            DAv2 -RS \cite{depthanythingv2} & MDE & Neurips24 & 0.869 & 0.937 & 0.109 & 0.480 & 0.063\\ 
            \midrule
            Ours-Small & DC & - & \cellcolor{yellow!25}0.921 & \cellcolor{yellow!25}0.963 & \cellcolor{yellow!25}0.080 & \cellcolor{yellow!25}0.379 & \cellcolor{yellow!25}0.038 \\
            Ours-Large & DC & - & \cellcolor{red!25}\textbf{0.933} & \cellcolor{red!25}\textbf{0.972} & \cellcolor{red!25}\textbf{0.075} & \cellcolor{red!25}\textbf{0.350} & \cellcolor{red!25}\textbf{0.034} \\
            \bottomrule
        \end{tabular}
    } 
    \vspace{-5pt}
    \caption{\textbf{Quantitative comparison on ZJU-L5}. MDE: Monocular Depth Estimation; DS: Depth Super-resolution; DC: Depth Completion. The \colorbox{red!25}{\textbf{best}} and \colorbox{yellow!25}{second best} are marked with colors, while the best result among existing methods is \underline{underlined}. CFPNet is the published SOTA method focusing on this dataset.}
    \label{tab:zjul5_quantitative}
    \vspace{-10pt}
\end{table}

\begin{table}[ht]
	\vspace{-6pt}
	\centering
	\renewcommand{\arraystretch}{1.5} 
	\setlength{\tabcolsep}{4.5pt} 
	\resizebox{\linewidth}{!}{ 
		\begin{tabular}{lcccccccc}
			\toprule
			\textbf{Method} & Deltar & CFPNet & CFormer & OMNI-DC & DAv2-L & DAv2-S & Ours-S & Ours-L \\
			\midrule
			Params \textit{(M)} & 18 & 20 & 81 & 84 & 335 & 24 & $6\,+\,\textit{24}$ & $12\,+\,\textit{24}$ \\
			FLOPs \textit{(G)}  & 42 & 46 & 380 & 398 & 674 & 47 & $26\,+\,\textit{13}$ & $64\,+\,\textit{47}$ \\
			\bottomrule
		\end{tabular}
	}
	\vspace{-5pt}
	\caption{\textbf{Complexity comparison.} We separately list the depth completion and the MDE model in our method, with FLOPs calculated at a resolution of 480$\times$640.}
	\label{tab:zjul5_complexity}
	\vspace{-10pt}
\end{table}

We found that due to changes in depth pattern and underlying anomalies, many models and designs that are effective in traditional benchmarks are not be well-suited for real-world dToF data. First, since depth points are roughly uniformly distributed yet extremely sparse (0.02\% in ZJU-L5 and 0.18\% in our data), 3D methods struggle to capture effective spatial interactions, and architectures sensitive to sparsity also suffer from performance degradation. Second, some designs that assume absolute accuracy of depth sensors are sensitive to real-world noise, focusing solely on preserving and rapidly propagating the sparse measurement. Examples include the point embedding operation in the affinity propagation module and residual connections with the initial sparse depth map.

\begin{figure}[htbp]
    \centering
    \includegraphics[width=\columnwidth]{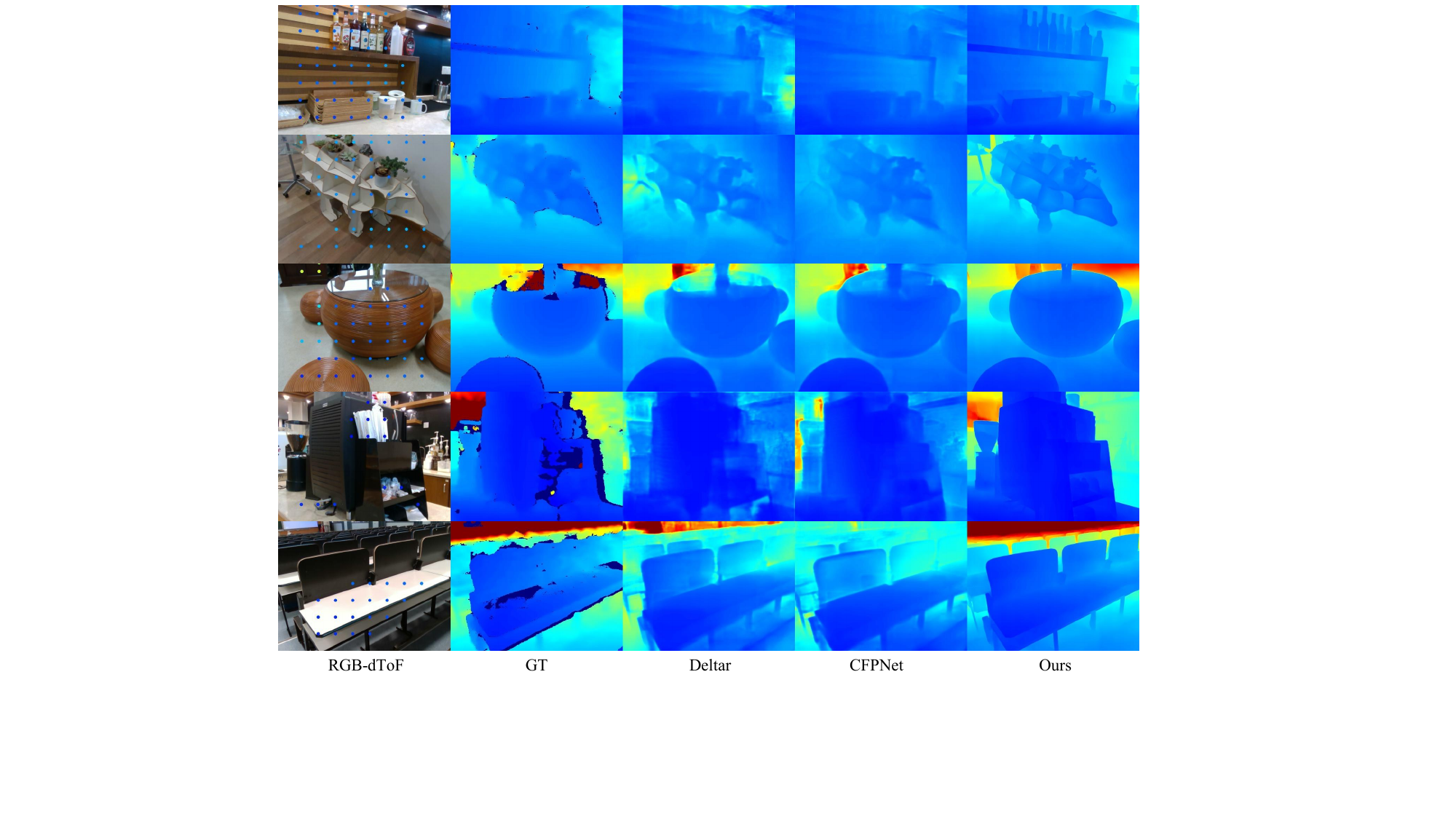}
    \vspace{-10pt}
    \caption{\textbf{Qualitative results on ZJU-L5}, our model further improve anomalies present in the ground truth}
    \label{fig:zju}
    \vspace{-10pt}
\end{figure}

In addition, we believe that the metrics from real datasets primarily reflect how well depth enhancement models improve low-cost sensors, bringing them closer to high-precision ones (specifically, L5 and RealScene 435i). However, even the depth sensors used to collect ground truth still exhibit limited performance in the challenging regions targeted in our study. Thus, these metrics may not fully capture the performance of our method. As shown in \cref{fig:zju}, our model's predictions not only outperform existing methods but also further improve anomalies present in the ground truth, which leads to a decrease in metrics. More qualitative results are provided in the supplementary material.

\noindent{\textbf{Results on our real-world data.}} For this type of data, we employed the full dToF simulation method described in \cref{sec:strategy} and maintained the same evaluation settings. Due to the higher image resolution, we modified some methods to accelerate training. 

\begin{table}[ht]
    \centering
    \renewcommand{\arraystretch}{1.1} 
    \setlength{\tabcolsep}{4pt} 
    \vspace{-6pt}
    \resizebox{\linewidth}{!}{%
        \begin{tabular}{lccccc}
            \toprule
            Model & $\delta_1$ & $\delta_2$ & RMSE & Rel & EWMAE \\
            \midrule
            BPNet\cite{bpnet} & - & - & 0.630 & - & - \\
            OMNI-DC\cite{omnidc} & 0.593 & 0.768 & 0.643 & 0.292 & 0.195 \\
            DAv2 -RS\cite{depthanythingv2} & 0.687 & 0.833 & \underline{0.292} & 0.237 & 0.141 \\
            PENet*\cite{penet} & \underline{0.740} & 0.878 & 0.327 & \underline{0.202} & \underline{0.139} \\ 
            CFormer*\cite{completionformer} & 0.732 & \underline{0.883} & 0.320 & 0.206 & 0.159 \\
            Ours & \textbf{0.790} & \textbf{0.911} & \textbf{0.226} & \textbf{0.155} & \textbf{0.108} \\
            \bottomrule
            \multicolumn{6}{l}{\footnotesize * indicates a lightweight version.}
        \end{tabular}
    }
    \vspace{-8pt}
    \caption{Quantitative comparison on our real-world samples.}
    \label{tab:sota}
    \vspace{-10pt}
\end{table}

\Cref{tab:sota} presents the quantitative results based on ground truth obtained from stereo matching. Despite some scale shifts and anomalies in the GT, these metrics still offer a meaningful preliminary evaluation of the model’s performance. Notably, our model achieves the best results.

\begin{figure}[htbp]
   \centering
   \includegraphics[width=\linewidth]{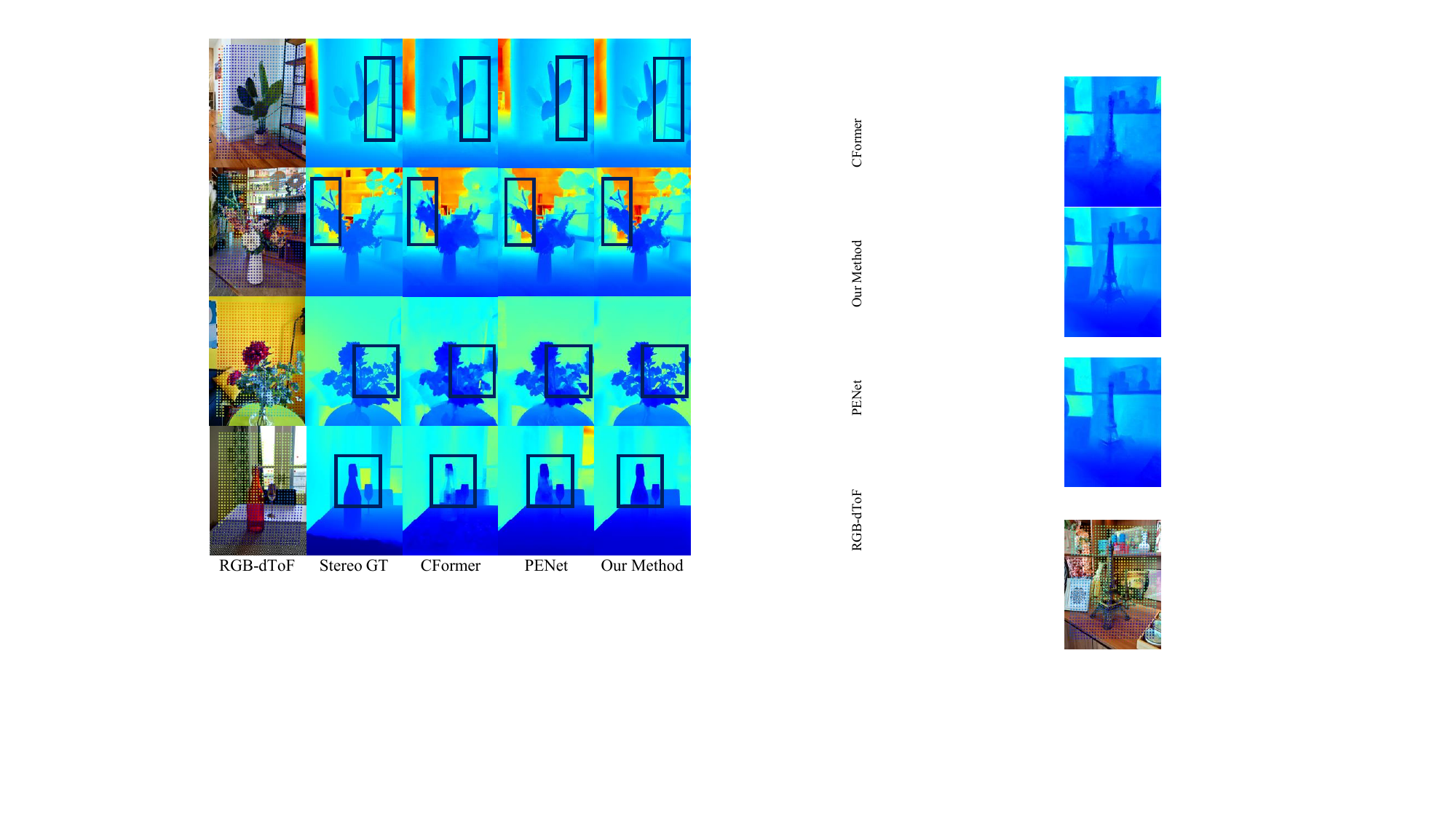}
   \vspace{-15pt}
   \caption{Qualitative results on our real-world samples.}
   \label{fig:compare}
   \vspace{-8pt}
\end{figure}

\Cref{fig:compare} presents the qualitative comparison. Our method effectively integrates the MDE model, improving performance in fine details and challenging regions. Additionally, we observed that PENet achieves better detail prediction than CFormer, which is consistent with the EWMAE metric in \Cref{tab:sota}. More qualitative results are provided in the supplementary material.

\subsection{Ablation Studies}
\label{sec:ablation}
\noindent \textbf{Components of Simulation Method.} We performed ablation studies on each component of our simulation method using the ZJU-L5 dataset, as shown in \cref{tab:ablationdegra}. Since calibration errors are not considered, we validate their impact through qualitative results on real-world samples, provided in the supplementary material.

\begin{table}[htbp]
    \vspace{-4pt}
    \centering
    \resizebox{\linewidth}{!}{%
        \begin{tabular}{lccccc}
            \toprule
            Method & $\delta_1$ & $\delta_2$ & Rel  & RMSE  & $\log_{10}$ \\
            \midrule
            Standard & \textbf{0.933} & \textbf{0.972} & \textbf{0.075} & \textbf{0.350} & \textbf{0.034}\\
            w/o RA & 0.923 & 0.966 & 0.076 & 0.362 & 0.037\\
            w/o OD & 0.912 & 0.965 & 0.091 & 0.381 & 0.043\\
            w/o SR & 0.773 & 0.847 & 0.175 & 0.566 & 0.138\\
            w/o (RA + OD) & 0.905 & 0.965 & 0.102 & 0.395 & 0.045\\  
            w/o (OD + SR) & 0.780 & 0.855 & 0.191 & 0.641 & 0.168\\  
            \bottomrule
        \end{tabular}
    }
    \vspace{-5pt}
    \caption{\textbf{Ablation studies about simulation method.} OD: Overall Distribution, SR: Specific Region, RA: Random Anomalies.}
    \label{tab:ablationdegra}
\end{table}

\noindent \textbf{Depthor with Different MDE Models.} As shown in \cref{tab:mde}, we replaced different MDE models in our method. 
We found that using more powerful MDE models does not significantly improve performance on ZJU-L5 compared to Hypersim, particularly in EWMAE, which further highlights the potential issues in the GT of real datasets.

\begin{table}[ht]
    \centering
    \vspace{-4pt}
    \resizebox{\linewidth}{!}{%
        \begin{tabular}{lcccccc}
            \toprule
            \multirow{3}{*}{Model} & \multicolumn{3}{c}{ZJU-L5} & \multicolumn{3}{c}{Hypersim} \\
            \cmidrule(lr){2-4} \cmidrule(lr){5-7}
            & RMSE & Rel & EWMAE & RMSE & Rel & EWMAE \\
            \midrule
            DAv2 -RS & 0.350 & 0.075 & 0.136 & 0.445 & 0.068 & 0.110 \\
            DAv2 -RB & 0.335 & 0.071 & 0.136 & 0.406 & 0.061 & 0.103 \\
            DAv2 -RL & 0.330 & 0.070 & 0.135 & 0.390 & 0.059 & 0.101 \\
            DAv2 -MS & 0.372 & 0.095 & 0.141 & 0.554 & 0.102 & 0.114 \\
            \bottomrule
        \end{tabular}
    }
    \vspace{-4pt}
    \caption{\textbf{Ablation studies on different MDE models}. The results on Hypersim are based on ZJU-L5's dToF simulation. R: relative; M: metric; S: small; B: base; L: large.}
    \label{tab:mde}
    \vspace{-8pt}
\end{table}

\begin{figure}[htbp]
	\centering
	\includegraphics[width=\linewidth]{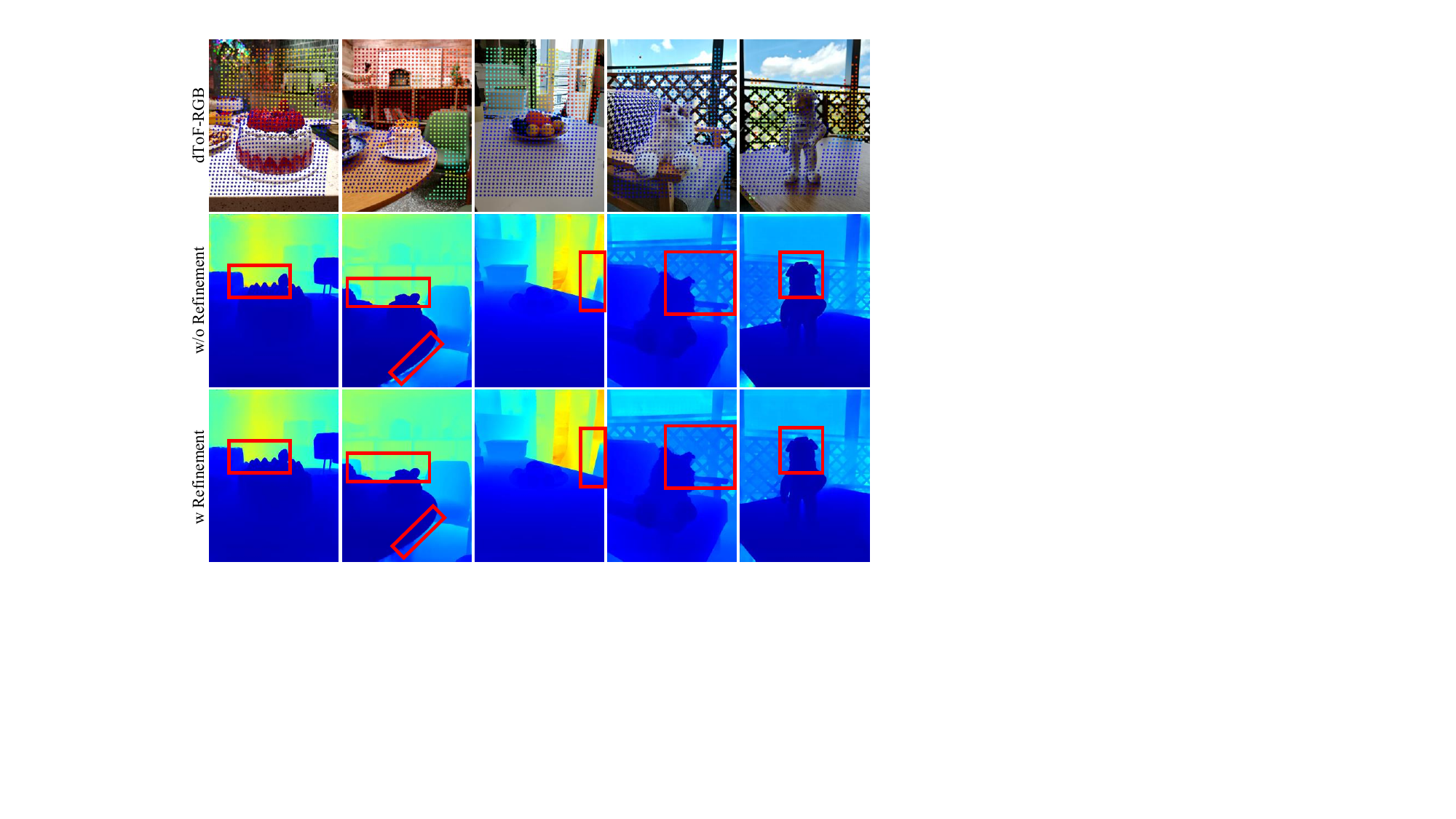}
	\vspace{-18pt}
	\caption{Refinement of mixed affinity propagation.}
	\label{fig:cspn}
	\vspace{-15pt}
\end{figure}

\noindent \textbf{Refinement of Mixed Affinity Propagation.} We analyze this module through quantitative metrics from synthetic datasets and qualitative results of real-world samples. As shown in \cref{tab:cspn}, on the Hypersim dataset, refining the initial depth map in full resolution using affinity propagation improves the model’s overall performance, particularly on boundary-focused metric EWMAE. The qualitative results in \cref{fig:cspn} also reveal that this module effectively improves the model's performance in regions beyond the sensor’s FoV and at foreground-background boundaries, mitigating anomalies while enhancing prediction consistency.

\begin{table}[ht]
    \centering
    \vspace{-4pt}
    \resizebox{\linewidth}{!}{%
        \begin{tabular}{ccccccc}
            \toprule
            \multirow{2}{*}{Refine} & \multicolumn{2}{c}{Input Feature} & \multicolumn{3}{c}{Hypersim} & \multirow{2}{*}{Params.} \\
            \cmidrule(lr){2-3} \cmidrule(lr){4-6}
            & MDE & UNet & RMSE & REL & EWMAE & (\textit{M}) \\
            \midrule
            / & / & / & 0.267 & 0.039 & 0.091 & - \\
            \checkmark & \checkmark & / & 0.269 & 0.039 & 0.087 & +0.048 \\
            \checkmark & / & \checkmark & 0.258 & 0.037 & 0.081 & +0.085 \\
            \checkmark & \checkmark & \checkmark & \textbf{0.248} & \textbf{0.034} & \textbf{0.079} & +0.122 \\
            \multicolumn{3}{c}{+ Point Embedding} & 0.328 & 0.038 & 0.098 & +0.122 \\
            \bottomrule
        \end{tabular}
    }
    \vspace{-6pt}
    \caption{\textbf{Ablation studies about refinement.} The results on Hypersim are based on our samples' dToF simulation}
    \label{tab:cspn}
    \vspace{-10pt}
\end{table}

Furthermore, our experimental results demonstrate that due to the lack of scale information and the resolution differences, computing affinity solely based on ${F}_{mde}$ improves EWMAE but adversely affects scale metrics. However, the contextual information in ${F}_{mde}$ can still be leveraged to enhance ${F}_{unet}$. Additionally, the regional characteristics and anomalies of dToF signals conflict with the assumptions of point embedding, leading to performance degradation.

\begin{figure}[htbp]
    \centering
    \vspace{-4pt}
    \includegraphics[width=\linewidth]{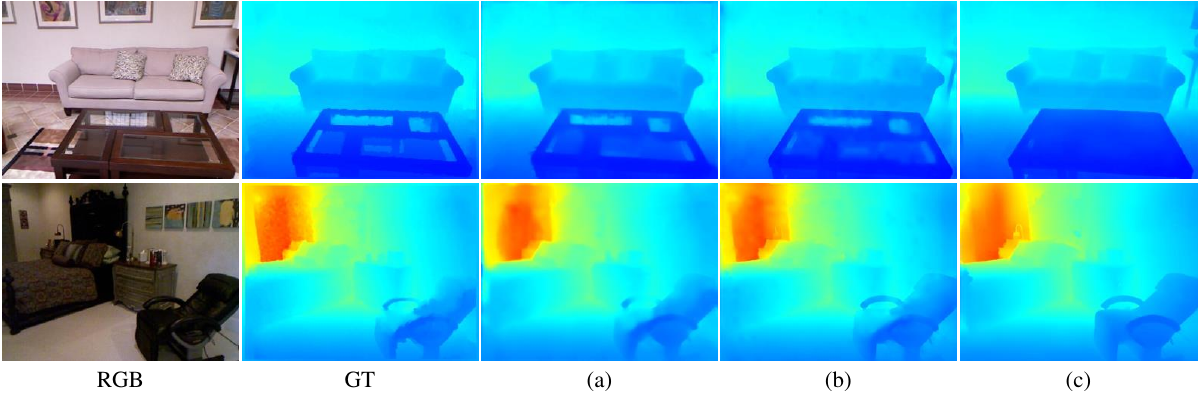}
    \vspace{-15pt}
    \caption{Prediction of our model under different combinations of training datasets and simulation methods.}
    \label{fig:complementarity}
    \vspace{-15pt}
\end{figure}

\noindent \textbf{Complementarity of Our Training Strategy and Model.} In \cref{fig:complementarity}, we present our model's predictions on the NYUv2 dataset under different training strategies, where dToF points are sampled from the inaccurate ground truth: \textit{(a)} Trained on the NYUv2 dataset, model tends to disregard MDE outputs due to conflicts with the inaccurate ground truth; \textit{(b)} Trained on the Hypersim dataset without our simulation method, model extracts only contextual information from MDE, to propagate accurate depth points while neglecting global depth relationships; \textit{(c)} Our training strategy enhances performance through both global relationships and local details.

%% file: sec/5_conclusion.tex
\section{Conclusion}
\label{sec:conclusion}

In this paper, we present a comprehensive solution to real-world dToF enhancement. Unlike previous super-resolution methods, we reformulate the problem within completion to streamline the task. We introduce a noise-robust training strategy that improves existing depth completion models, achieving results comparable to depth super-resolution methods. Additionally, we design a novel network that effectively integrates MDE to enhance predictions in challenging regions. Our method achieves state-of-the-art results on both the ZJU-L5 dataset and a challenging set of dToF samples. Not fast enough for real-time inference is the main limitation of our method. Extending our method to other sensors is a promising direction for future research.

\begin{sloppypar}
	\noindent\textbf{Acknowledgement.} This research is supported by the National Key R\&D Program of China (2024YFE0217700), National Natural Science Foundation of China (62472184), the Fundamental Research Funds for the Central Universities, and the Innovation Project of Optics Valley Laboratory (Grant No. OVL2025YZ005)
\end{sloppypar}

%% file: sec/X_suppl.tex
\clearpage
\setcounter{page}{1}
\maketitlesupplementary
This supplementary material provides additional details to complement the main paper. It includes introduction of dToF imaging (\cref{sec:preliminary}), detailed training settings (\cref{sec:supsetting}), descriptions of the adopted evaluation metrics (\cref{sec:supmetric}),  introduction of dToF projection (\cref{sec:supproject}), implementation details of the dToF simulation method (\cref{sec:supdetail}), additional ablation studies about simulation method (\cref{sec:supdegradation}), and additional experimental results (\cref{sec:supresult}).

\section{Preliminary: dToF Imaging}
\label{sec:preliminary}
\begin{figure}[htbp]
	\centering
	\includegraphics[width=0.9\linewidth]{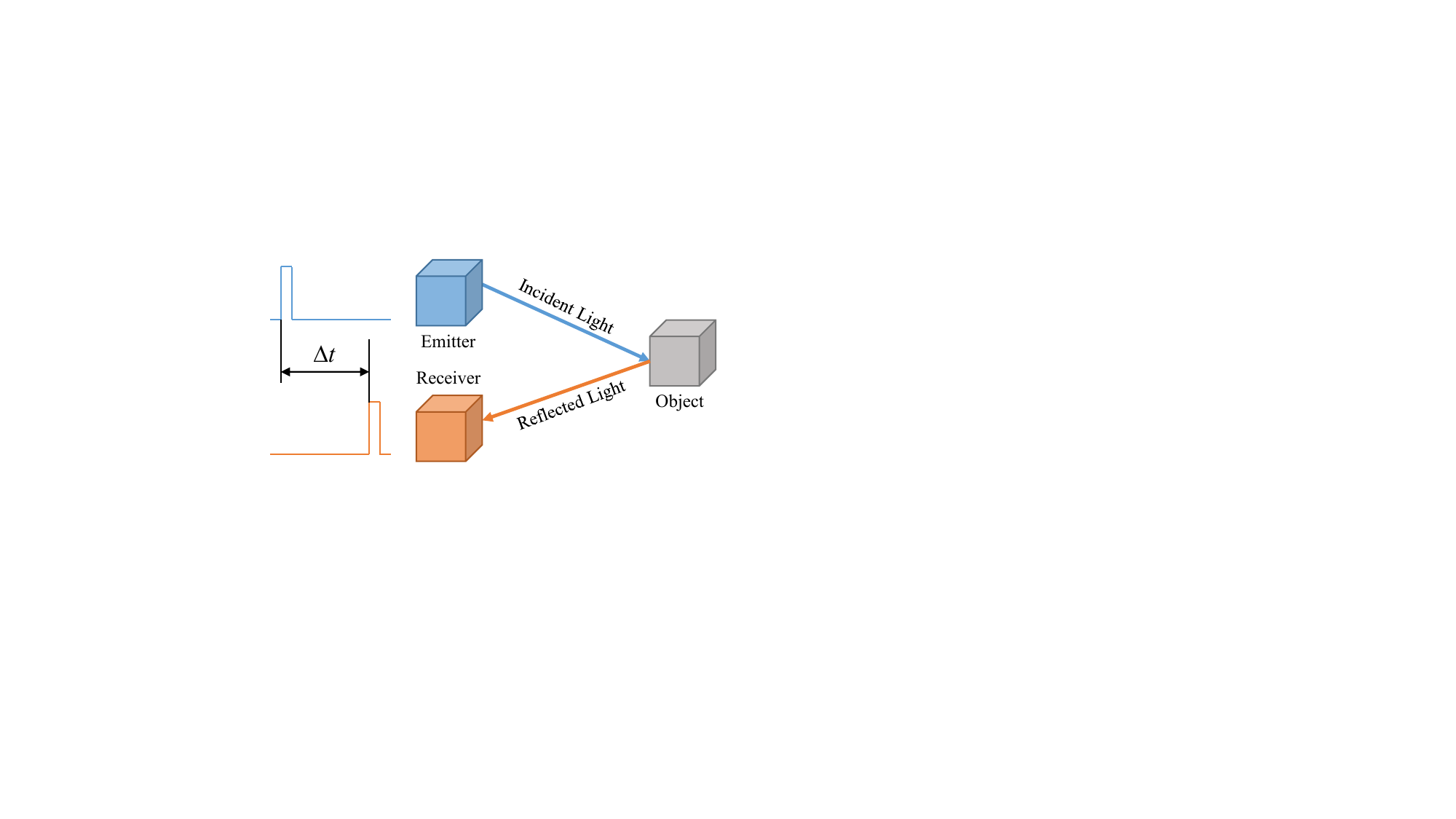}
	\caption{Imaging principle of direct Time-of-Flight sensor}
	\label{fig:principle}
	\vspace{-1em} 
\end{figure}

We first briefly introduce the imaging principle of dToF. As shown in \cref{fig:principle}, a pulsed laser generates a short light pulse and emits it into the scene. The pulse scatters, and some photons are reflected back to the dToF detector. The depth is then determined by the formula $d=\Delta t\cdot c/2$, where $\Delta t$ is the time difference between laser emission and reception, and $c$ is the speed of light. Each dToF pixel captures all scene points reflected within its individual field-of-view (iFoV) using time-correlated single-photon counting (TCSPC). The iFoV is determined by the sensor’s total field-of-view (FoV) and spatial resolution, returning the peak signal detected within that range. Interested readers are referred to \cite{dvsr,dtof0,dtof1} for more details.

\section{Training Setting.}
\label{sec:supsetting}
We implement our method in pytorch\cite{paszke2019pytorch} and train it on 4 Nvidia RTX 3090 GPUs. We adopt AdamW\cite{adamw} with 0.1 weight decay as the optimizer, and clip gradient whose ${{l}^{2}}$-norm is larger than 0.1. Our model is trained from scratch in roughly 230K iterations using the OneCycle\cite{onecycle} learning rate policy, setting the initial learning rate to 1/25 of the maximum learning rate and gradually reducing the learning rate to 1/100 of the maximum learning rate in the later stages of training. We set batch size as 12 and the largest learning rate as 0.0003.
\section{Details on Evaluation Metrics}
\label{sec:supmetric}
We present the precise definitions of the quantitative metrics reported in the main paper, which include ${\delta}_{i}$, Rel, RMSE, $\log_{10}$, and edge-weighted mean absolute error (EWMAE). These metrics are defined as follows:
\[
\begin{aligned}
  & \text{Rel} = \frac{1}{|P|} \sum \frac{|y_p - x_p|}{y_p}, \\ 
  & \text{RMSE} = \sqrt{\frac{1}{|P|} \sum (y_p - x_p)^2}, \\ 
  & \text{EWMAE} = \frac{1}{|P|} \frac{\sum G_p \cdot |y_p - x_p|}{\sum G_p},\\
  & {\delta}_{i} = \frac{1}{|P|} \sum \left( \max \left( \frac{y_p}{x_p}, \frac{x_p}{y_p} \right) < 1.25^i \right), \\ 
  &\log_{10} = \frac{1}{|P|} \sum \left| \log_{10} (y_p) - \log_{10} (x_p) \right|
\end{aligned}
\]

Here, ${{x}_{p}}$ and ${{y}_{p}}$ represent the predicted value and ground truth at valid pixel locations, respectively. The set ${P}$ contains all pixels with valid ground truth, and ${|P|}$ denotes the total number of such pixels.

Following \cite{mipi2022,mipi2023,ewmae}, we compute the weight coefficient $G_p$ for a pixel $p$ based on its intensity and directional gradients. First, the directional gradient ${{\nabla }_{D}}I(p)$ is calculated as: \[{{\nabla }_{D}}I(p)={{V}_{pD}}-{{V}_{p}}\]

where $D\in \{N,S,E,W\}$ represents the north, south, east, and west neighbors of pixel $p$ and 
$V_p$ is the depth of $p$. Using these gradients, we compute the reciprocals of directional conduction functions ${{G}_{{{D}_{p}}}}$, which is expressed as: \[{{G}_{{{D}_{p}}}}=\frac{{{\left[ {{\nabla }_{D}}I(p) \right]}^{2}}}{{{\left[ {{\nabla }_{D}}I(p) \right]}^{2}}+{{\kappa }^{2}}}\]

$\kappa$ is a regularization constant. Finally, the weight coefficient $G_p$ is obtained as the average of these directional coefficients: \[{{G}_{p}}=\frac{{{G}_{{{N}_{p}}}}+{{G}_{{{S}_{p}}}}+{{G}_{{{E}_{p}}}}+{{G}_{{{E}_{p}}}}}{4}\]

Each pixel's weight ${G}_{p}$ can be calculated based on the above formula. The weight approaches 0 when the pixel is in a homogeneous region and approaches 1 when the gradient in all four directions reaches a maximum.

\section{Project dToF to Sparse Depth Map}
\label{sec:supproject}

Each dToF measurement provides a 3D point in the dToF sensor coordinate system:

\begin{equation}
P_{dToF} = (X_{dToF}, Y_{dToF}, Z_{dToF}, 1)^T.
\end{equation}

The transformation from the dToF coordinate system to the RGB camera coordinate system is given by:

\begin{equation}
P_{RGB} = T_{dToF \rightarrow RGB} P_{dToF},
\end{equation}

where the transformation matrix is:

\begin{equation}
\begin{aligned}
    &T_{dToF \rightarrow RGB} =
    \begin{bmatrix} 
        R_T & t_T \\ 
        0 & 1 
    \end{bmatrix}, \\
    &R_T = R_{RGB} R_{dToF}^{-1}, \\
    &t_T = t_{RGB} - R_{RGB} R_{dToF}^{-1} t_{dToF}.
\end{aligned}
\end{equation}

The transformed 3D point is then projected onto the RGB image using the intrinsic matrix \( K_{RGB} \) to get the homogeneous image coordinates:

\begin{equation}
\begin{bmatrix} u \\ v \\ w \end{bmatrix} =
K_{RGB} 
\begin{bmatrix} X_{RGB} \\ Y_{RGB} \\ Z_{RGB} \end{bmatrix}.
\end{equation}

where

\begin{equation}
K_{RGB} =
\begin{bmatrix} 
f_x & 0 & c_x \\ 
0 & f_y & c_y \\ 
0 & 0 & 1 
\end{bmatrix}.
\end{equation}

The final pixel coordinates \((u, v)\) are obtained via perspective division:

\begin{equation}
u = \frac{f_x X_{RGB}}{Z_{RGB}} + c_x, \quad
v = \frac{f_y Y_{RGB}}{Z_{RGB}} + c_y.
\end{equation}

Existing depth super-resolution methods typically compute the iFoV region coordinates for each measurement based on this central coordinate, resolution, and FoV. However, calibration errors can cause significant shifts in these depth points. Therefore, we approach this problem from the perspective of depth completion robustness.

\section{Details of dToF Simulation Method}
\label{sec:supdetail}
We trained our model on the Hypersim dataset. To reduce the impact of invalid data, we scaled some of the depth values that exceeded the sensor’s detection limit. Similar to the approach of Sun \etal~\cite{dvsr} on \cite{tartanair}, if 60\% or more of the depth values in an image exceed 6 meters, all depth values are halved. Additionally, we modified the parameters of our simulation method for each test dataset to match the characteristics of different dToF sensors.

\noindent{\textbf{ZJU-L5 Dataset.}} The resolution of the dToF sensor and the depth ground truth are $8\times8$ and $480\times640$, respectively. According to the calibration results provided by the authors, the FoV of the L5 sensor covers approximately 61\% of the GT. The mean boundary values of its projected region on the GT are [-25, 405, 85, 535], corresponding to the upper ($h_u$), lower ($h_l$), left ($w_l$), and right ($w_r$) boundaries, respectively. Each dToF signal corresponds to an iFoV of approximately $52 \times 56$ pixels. Additionally, the maximum depth recorded by the L5 sensor is $4.1$ m, whereas the maximum depth in the GT is $10$ m.

Due to the low power of the L5 sensor, it typically exhibits signal loss in specific regions rather than returning incorrect depth values. Based on the dataset masks, the probability of signal loss is approximately 30\%.  As the authors performed strict calibration and no noticeable calibration errors were observed in the visualization results, we did not consider region shift in our simulation method.

\noindent{\textbf{Our Real-world Samples.}} The resolution of the dToF sensor and RGB camera are $40\times30$ and $912\times684$, respectively. To allow for 1/32 downsampling, we padded the images to $928\times714$. We used the internal parameters of the mobile phone to project the raw dToF signals; the FoV of the dToF sensor covers approximately 81\% of the image. The mean boundary values of its projected region on the image are [30, 900, 40, 660]. Each dToF signal corresponds to an iFoV of approximately $21 \times 21$ pixels. Additionally, the maximum depth recorded by the dToF sensor is $6$ m, whereas the theoretical detection limit is $8.1$ m.

Due to the higher performance of the dToF sensor, it can still receive photons that pass through non-Lambertian surfaces and may return valid depth values even in low-reflectivity regions. As a result, the collected samples exhibit more complex anomalies. To address this, we set the probability of depth loss to 80\% for pixels with a V-channel value below 40 in the HSV color space and assigned corresponding anomalies based on semantic labels.

\section{Ablation Studies of dToF Simulation}
\label{sec:supdegradation}

We demonstrated the effectiveness of certain components of our simulation method through quantitative results on the ZJU-L5 dataset in the main paper. Since the calibartion errors are not considered on the ZJU-L5, in this section, we provide additional visualizations on our collected data as a supplement. The results presented were obtained by training the lightweight PENet \cite{penet} on the Hypersim \cite{hypersim} dataset and evaluating its performance on real-world data. 

During these experiments, we simulated signal loss in distant regions and applied supervision, which occasionally caused the model to predict areas with missing depth input as distant regions incorrectly, since PENet lack of global relationships provided by the MDE model.

\begin{figure}[htbp]
    \centering
    \includegraphics[width=\linewidth]{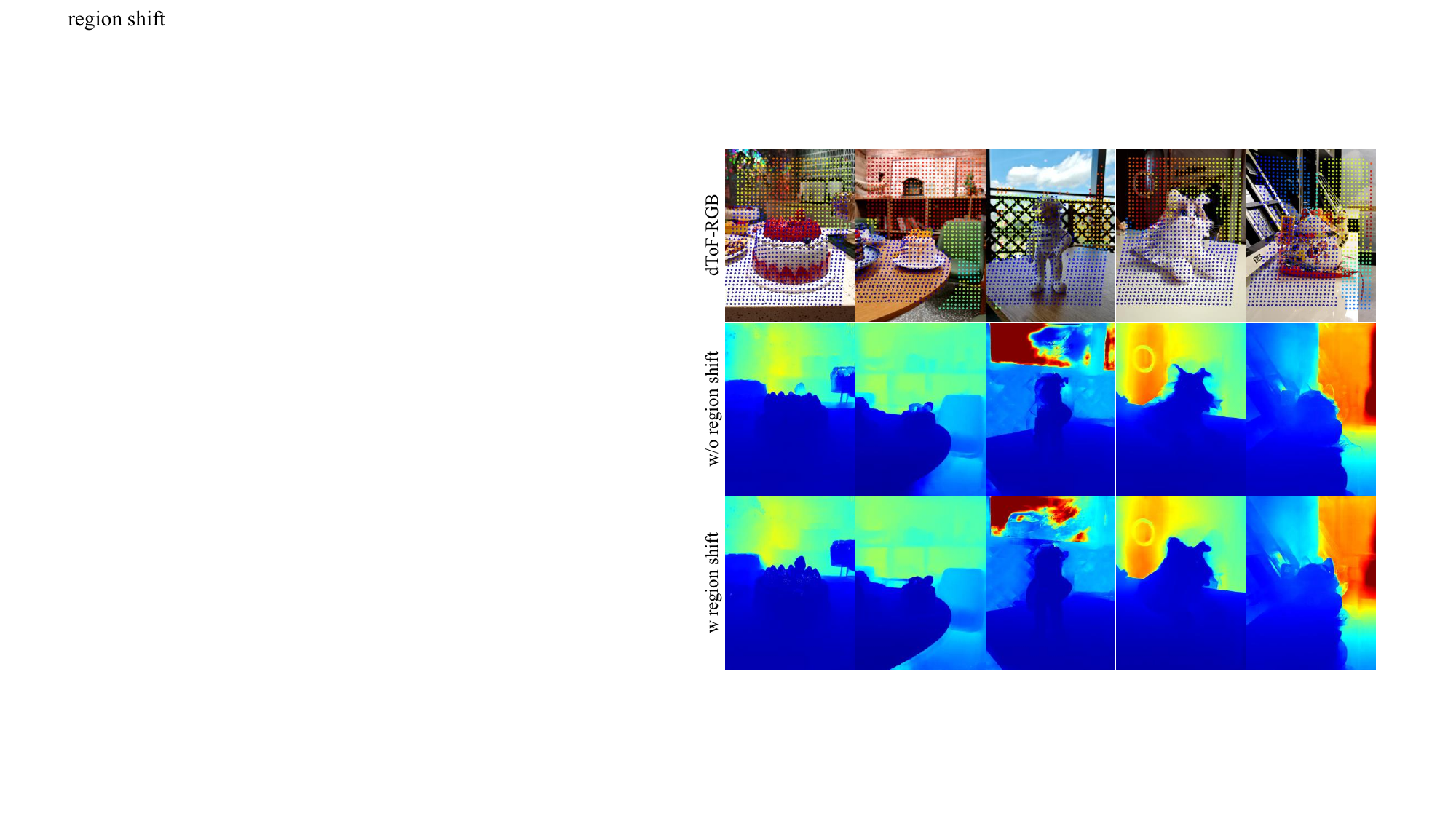}
    \caption{{\textbf{Effect of simulating calibration errors.} Prediction results are generated by the lightweight PENet\cite{penet}.}}
    \label{fig:regionshift}
\end{figure}

\Cref{fig:regionshift} illustrates the improvements in boundary predictions achieved by incorporating region shifts. These include resolving foreground-background overlaps caused by calibration errors and correcting errors at object boundaries, where dToF depth points represent the regional peak value.

\begin{figure}[htbp]
    \centering
    \includegraphics[width=\linewidth]{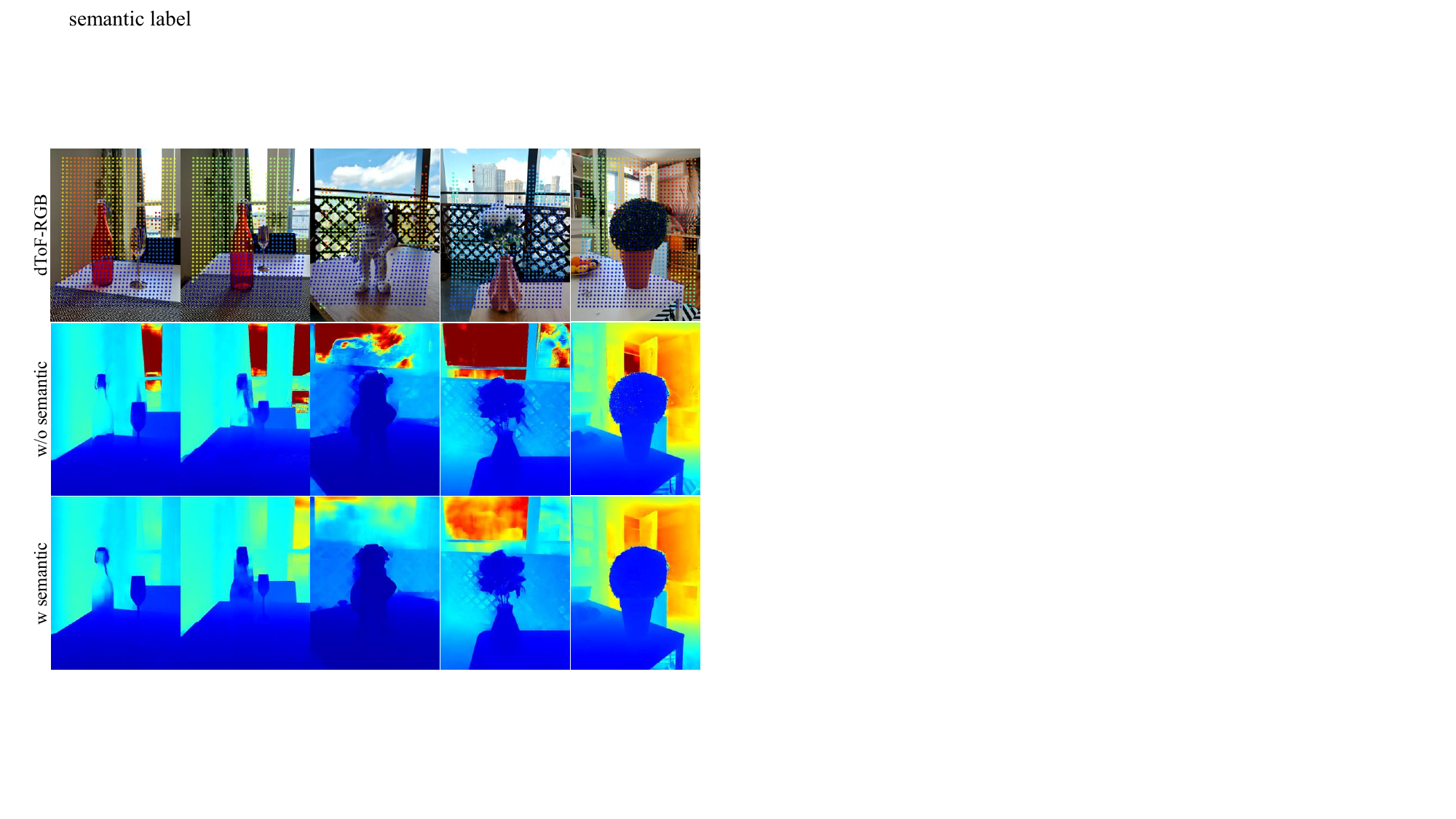}
    \caption{\textbf{Effect of simulating Non-Lambertian regions.} Prediction results are generated by the lightweight PENet\cite{penet}.}
    \label{fig:semantic}
    \vspace{-1em} 
\end{figure}

\Cref{fig:semantic} illustrates the results of simulating non-Lambertian surfaces. In cases of signal loss, the model utilizes surrounding information to predict values instead of directly assigning distant depths. Moreover, when photons pass through objects and return erroneous values, the model demonstrates the ability to partially correct these signals.

\section{Additional Experimental Results}
\label{sec:supresult}
\begin{figure*}[tbp]
    \setlength{\abovecaptionskip}{5pt}  
    \setlength{\belowcaptionskip}{5pt}  
    \centering
    \includegraphics[width=0.95\linewidth]{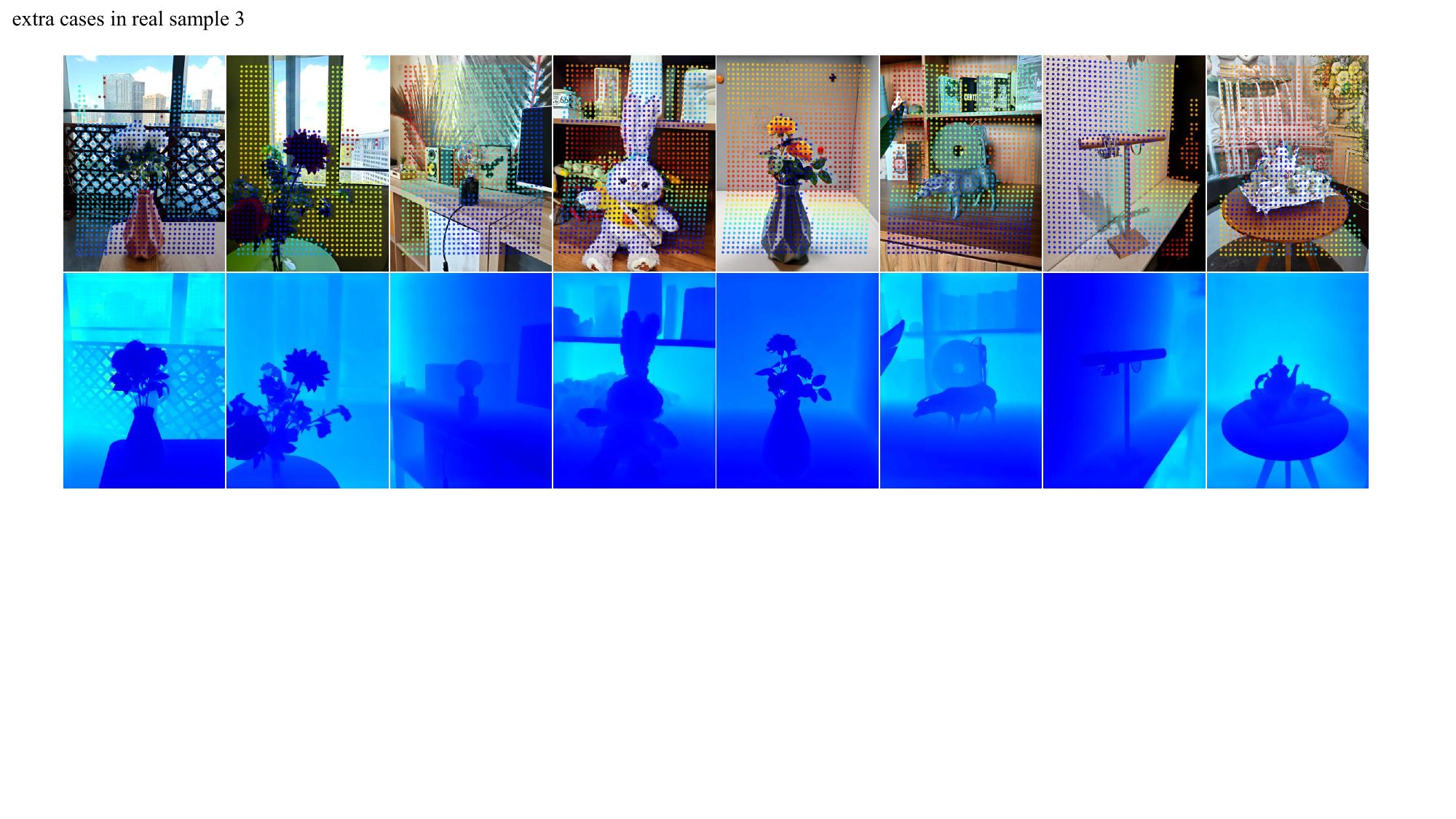}
    \includegraphics[width=0.95\linewidth]{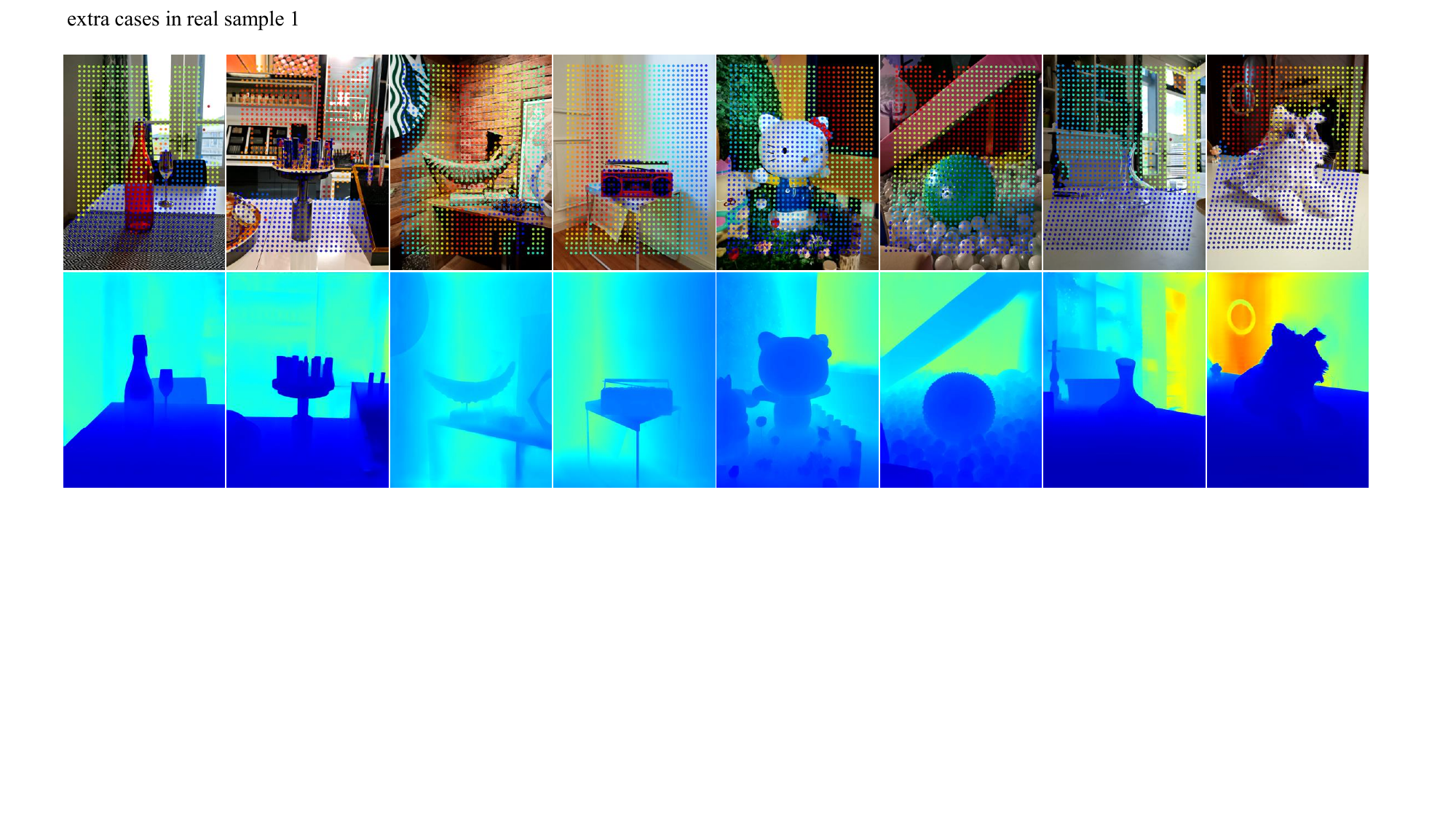}
    \includegraphics[width=0.95\linewidth]{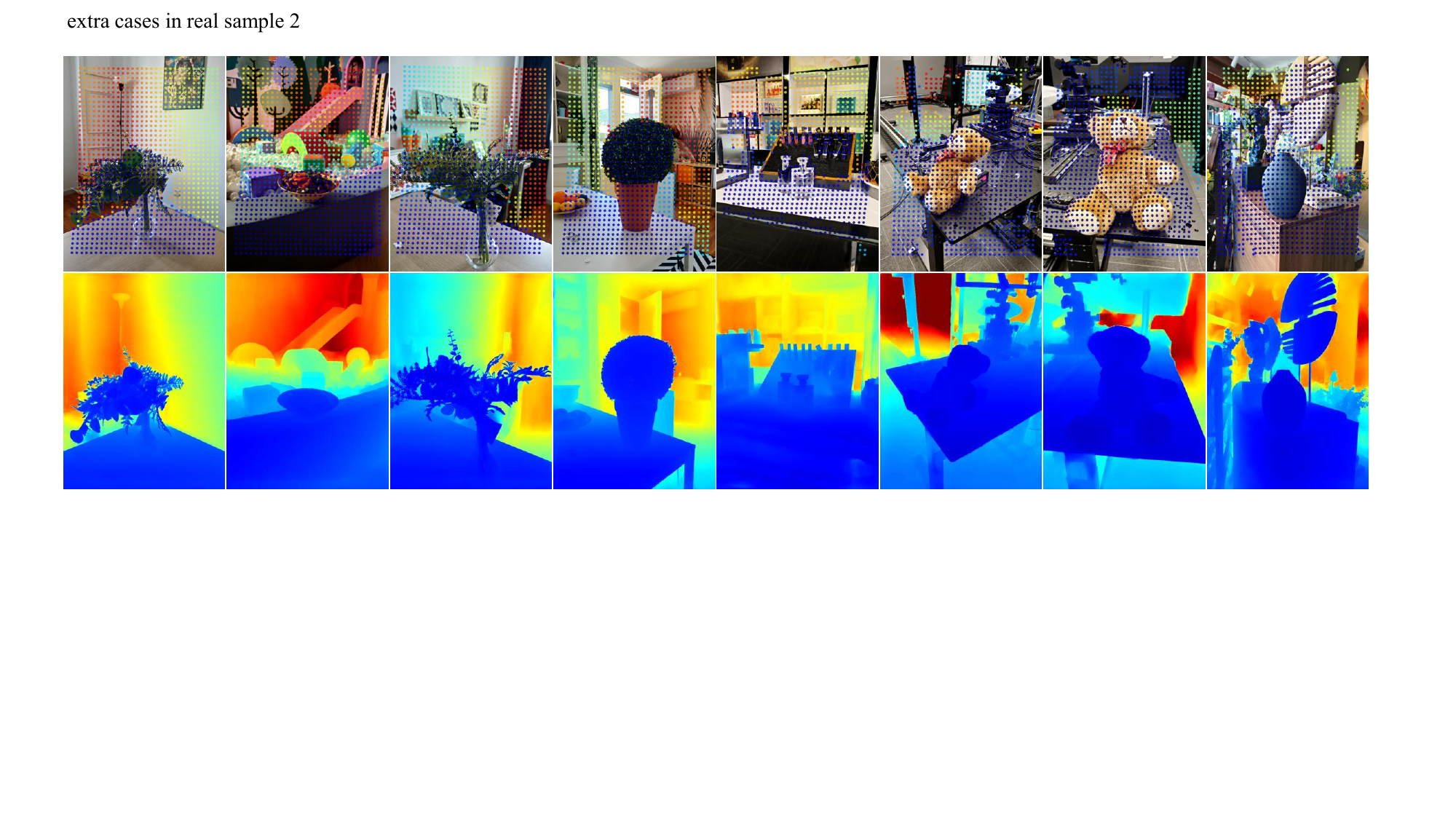}
    \caption{Additional qualitative results on real-world dToF samples.}
    \label{fig:extradtof}
    \vspace{-1em} 
\end{figure*}

\begin{figure*}[bp]
    \setlength{\abovecaptionskip}{5pt}  
    \centering
    \includegraphics[width=0.95\linewidth]{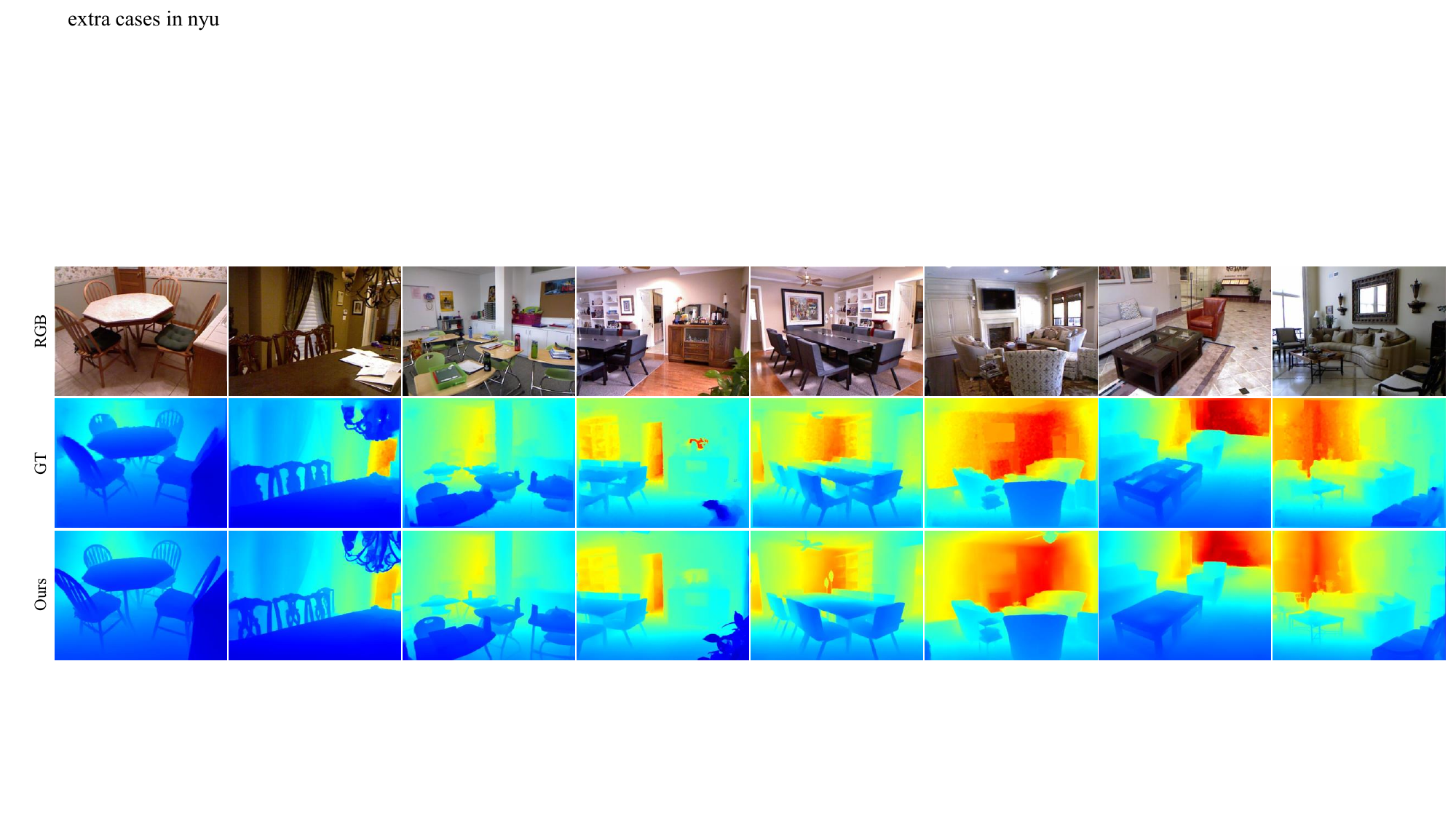}
    \caption{Additional qualitative results on NYUv2 dataset. From top to bottom: RGB, GT, Our results}
    \label{fig:extranyu}
\end{figure*}

\begin{figure*}[tbp]
    \setlength{\abovecaptionskip}{5pt}  
    \setlength{\belowcaptionskip}{5pt}  
    \centering
    \includegraphics[width=0.8\linewidth]{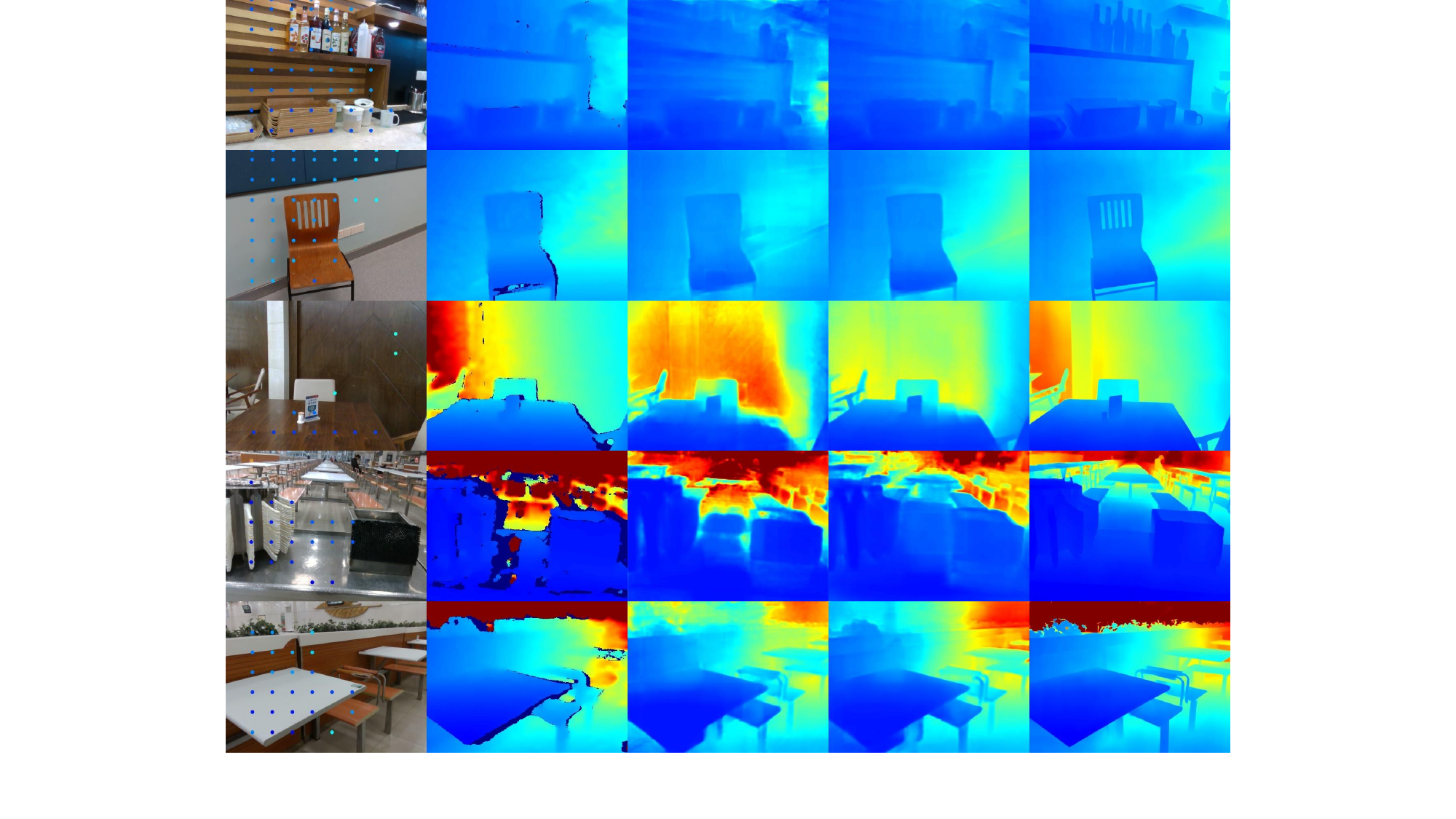}
    \includegraphics[width=0.8\linewidth]{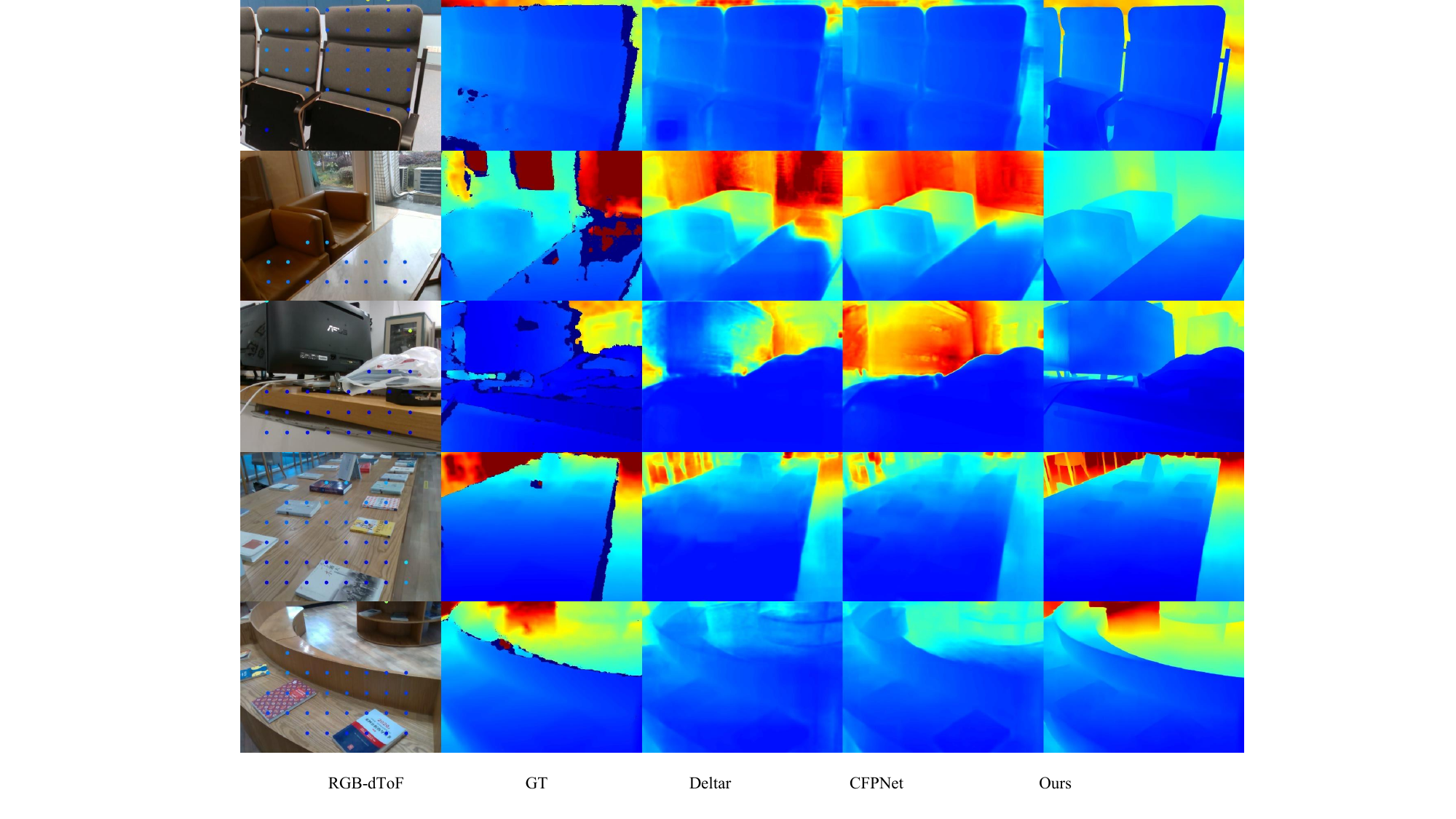}
    \caption{Additional qualitative results on ZJU-L5. From left to right, RGB-dToF, GT, Deltar, CFPNet, Our results.}
    \label{fig:extradzju}
    \vspace{-1em} 
\end{figure*}

Due to space limitations, we present additional experimental results here. \Cref{fig:extradtof}, \cref{fig:extradzju} and \cref{fig:extranyu} show the results on our dToF samples, the ZJU-L5 dataset and the NYUv2 dataset, respectively.

\begin{figure}[hbtp]
    \centering
    \includegraphics[width=\linewidth]{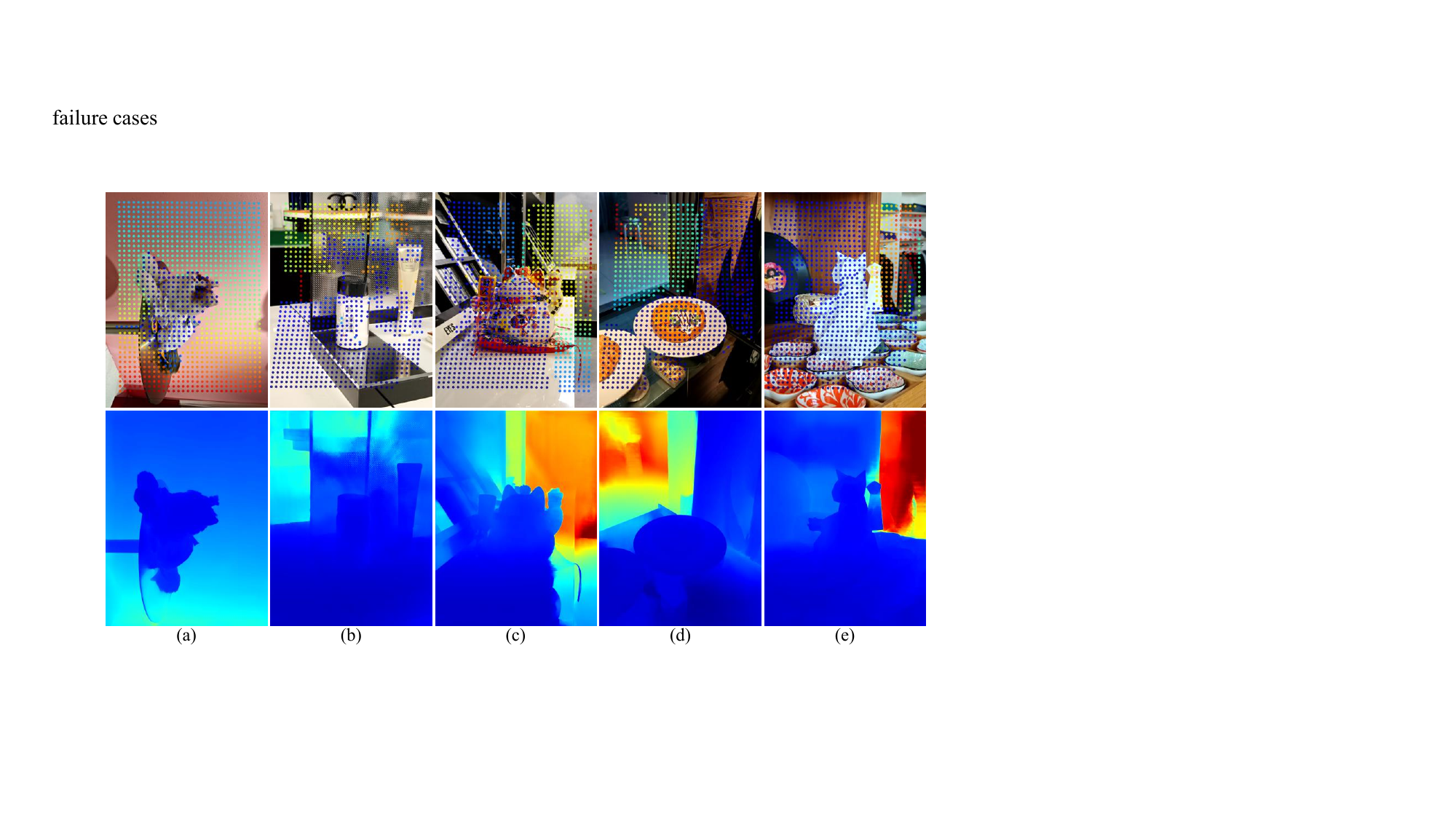}
    \caption{Failure example of real dToF data.}
    \label{fig:extrafailure}
\end{figure}

\Cref{fig:extrafailure}  presents failure cases from real dToF data, primarily caused by excessive dToF anomalies, while our model shows some improvement in handling these issues, such as correcting the sculpture's arm in \cref{fig:extrafailure}e and \cref{fig:extrafailure}c, further refinement is needed. Additionally, the MDE model exhibited semantic errors when processing rotated images, failing to correct the anomaly in \cref{fig:extrafailure}a. This issue can be resolved by converting the images to a normal perspective.
